\documentclass{article} % For LaTeX2e
\usepackage{iclr2027_conference,times}

\usepackage{amsmath,amsfonts,bm}

\def\eqref#1{equation~\ref{#1}}
\def\1{\bm{1}}

\DeclareMathAlphabet{\mathsfit}{\encodingdefault}{\sfdefault}{m}{sl}
\SetMathAlphabet{\mathsfit}{bold}{\encodingdefault}{\sfdefault}{bx}{n}

\usepackage{xurl}
\usepackage{hyperref}
\usepackage{url}
\usepackage{booktabs}
\usepackage{graphicx}
\usepackage{wrapfig}
\usepackage{capt-of}
\usepackage{subcaption}
\usepackage{xcolor}
\definecolor{bestcolor}{RGB}{220,20,60}
\definecolor{secondcolor}{RGB}{145,30,180}
\definecolor{thirdcolor}{RGB}{0,90,200}
\newcommand{\best}[1]{\textcolor{bestcolor}{\textbf{#1}}}
\newcommand{\second}[1]{\textcolor{secondcolor}{\underline{#1}}}
\newcommand{\third}[1]{\textcolor{thirdcolor}{#1}}
\newcommand{\dropval}[1]{\textcolor{red}{-$#1$}}

\makeatletter
\newcommand{\finishwrap}{%
    \par
    \ifx\parshape\WF@fudgeparshape
        \nobreak
        \ifnum\c@WF@wrappedlines>\@ne
            \advance\c@WF@wrappedlines\m@ne
            \vskip\c@WF@wrappedlines\baselineskip
            \global\c@WF@wrappedlines\z@
        \fi
        \allowbreak
        \WF@finale
    \fi
    \leavevmode
}
\makeatother

\title{MambaTG: Candidate-Aware State Space Modeling for Future Link Prediction on Temporal Graphs}

\author{%
Nguyen Hong Quan$^{1}$,\quad
Sitao Luan$^{2,3\,}$\thanks{Corresponding author. Email address:
\nolinkurl{nguyenho003@e.ntu.edu.sg},
\nolinkurl{sitao.luan@mail.mcgill.ca},
\nolinkurl{yilun001@e.ntu.edu.sg},
\nolinkurl{elhchen@ntu.edu.sg}.},\quad
Yilun Zheng$^{1}$,\quad
Lihui Chen$^{1\,*}$ \\
\normalfont
$^{1}$Nanyang Technological University, Singapore \\
$^{2}$University of Montreal\\
$^{3}$Mila -- Quebec Artificial Intelligence Institute
}

\iclrfinalcopy % Uncomment for camera-ready version, but NOT for submission.
\begin{document}

\maketitle
\lhead{Preprint}

\begin{abstract}

Future link prediction on temporal graphs involves both repeated interactions between previously connected node pairs and first-time interactions between pairs with no prior interaction. While repeated-edge prediction can exploit direct pairwise evidence, unseen-edge prediction must generalize beyond historical recurrence. Existing temporal graph models often exploit signals tied to recurring interactions or historical overlap, which are highly informative for repeated edges but become limited or unavailable for unseen edges, contributing to substantial performance degradation on unseen-edge prediction. Based on this observation, we identify two complementary signals for modeling sequential dynamics beyond historical recurrence: (1) \emph{historical-sequence evolution}, which captures dependencies among the source node's chronologically ordered interactions, and (2) \emph{candidate--history compatibility}, which captures how well a candidate fits the source node's evolving history. Based on the decomposition, we propose \textbf{MambaTG}, a candidate-aware Mamba-based framework that jointly models these two signals. Mamba provides a natural mechanism for this design: its selective recurrence captures the evolution of ordered interaction histories, while its input-dependent state transitions allow the history state to adapt to different candidate inputs. Extensive experiments on 13 temporal graph datasets against 13 representative baselines show that MambaTG achieves strong and consistent performance across both repeated and unseen edges while remaining computationally efficient.

\end{abstract}
\section{Introduction}
Temporal graphs represent entities as nodes and their interactions as time-stamped edges \citep{DBLP:journals/corr/abs-1905-11485}. They provide a natural framework for modeling evolving relationships in applications such as social networks \citep{Alvarez_Rodriguez_2021} and recommender systems \citep{yu2023continuoustimeuserpreferencemodelling}. A fundamental task in temporal graph learning is \emph{future link prediction}, which predicts interactions at a future query time from historical interactions \citep{10.1007/s00354-019-00065-z}. Future interactions can be divided into two categories, (1) \emph{repeated edges}, connecting previously interacting node pairs, and (2) \emph{unseen edges}, connecting pairs with no prior interaction \citep{yi2025tgbseqbenchmarkchallengingtemporal}. Both arise naturally in real-world systems: users on a social platform may repeatedly interact with existing connections while also forming new connections with previously unrelated users. Despite their coexistence, the two regimes present distinct predictive challenges. Repeated-edge prediction can directly exploit prior pairwise interactions, whereas unseen-edge prediction must infer future relationships from broader patterns in historical behaviors. A practical temporal graph model must therefore capture recurring relationships while also generalizing to new ones without prior pairwise interactions.

However, some existing temporal link prediction methods still rely heavily on signals associated with repeated interactions. The competitiveness of EdgeBank, a non-parametric baseline that predicts links by memorizing previously observed edges, shows that historical recurrence alone can provide a remarkably strong signal on conventional benchmarks \citep{poursafaei2022towards}. This bias toward recurrence is reflected in representative temporal graph neural network (T-GNN) designs. Recent methods such as DyGFormer\citep{dygformer}, FreeDyG\citep{tian2024freedyg}, DyG-Mamba\citep{li2025dyg-mambacontinuousstatespace}, and DyGMamba\citep{ding2025dygmambaefficientlymodelinglongterm} explicitly encode source--destination historical overlap through neighbor co-occurrence or interaction-frequency features. Such signals are highly informative for repeated edges but may be sparse or unavailable for unseen edges, particularly in bipartite graphs. Consistent with this limitation, recent benchmarks report substantially lower performance on unseen edges than on repeated edges \citep{yi2025tgbseqbenchmarkchallengingtemporal}. These observations suggest that effective future link prediction requires moving beyond recurrence-based evidence and learning historical patterns that can generalize to previously unseen interactions.

\begin{figure}[t]
    \centering
    \includegraphics[width=\linewidth]{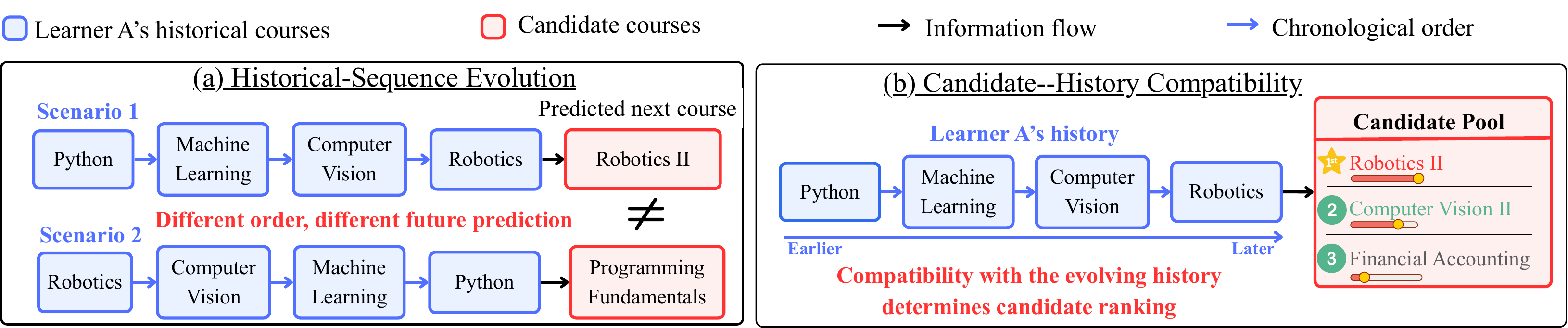}
    \caption{Illustration of two complementary signals for predicting Learner A's next course choice from previously taken courses in Python, Machine Learning, Computer Vision, and Robotics. }
    \label{fig:two-signals}
    \vspace{-8pt}
\end{figure}

This raises a more fundamental question: \emph{\textbf{What predictive information should a temporal graph model learn from historical interactions to generalize beyond direct pairwise recurrence?}} We argue that the answer lies in two complementary signals: \emph{historical-sequence evolution} and \emph{candidate--history compatibility}, as illustrated through Learner A's historical course interactions in Figure~\ref{fig:two-signals}. \textbf{(1) Historical-sequence evolution} (Figure~\ref{fig:two-signals}(a)) captures how historical interactions evolve in temporal order. The same interactions can imply different future behavior when ordered differently, as later interactions may reflect intentions developed through earlier ones. Capturing this signal therefore requires modeling dependencies among historical interactions as an ordered sequence, such that later interactions are interpreted in the context of the preceding history. \textbf{(2) Candidate--history compatibility} (Figure~\ref{fig:two-signals}(b)) captures how well a candidate fits the source node's evolving history. Explicitly modeling this relationship allows predictions to adapt to each candidate. However, many T-GNNs construct source and candidate representations largely independently and combine them only at the prediction stage \citep{jodie,graphmixer}, limiting direct modeling of the relationship between the candidate and the source history. CRAFT~\citep{craft} directly matches each candidate against the source node's individual historical interactions, but does not explicitly model dependencies arising from the history's sequential evolution before candidate matching. Overall, the two signals capture distinct but complementary aspects of future interaction prediction: \emph{historical-sequence evolution} characterizes dependencies within the ordered interaction history, whereas \emph{candidate--history compatibility} reflects how that evolving history informs the prediction of a candidate interaction.

Jointly modeling these complementary signals poses a non-trivial design challenge: historical-sequence evolution is shared across candidates, whereas candidate--history compatibility is inherently candidate-specific. To address this challenge, we propose MambaTG, an SSM-based framework that uses Mamba~\citep{gu2024mambalineartimesequencemodeling} as its sequential encoder. Its Candidate-Aware Mamba Encoder (CAME) first encodes the source's ordered historical interactions into a shared history state and then adapts this state to each candidate through Mamba's input-dependent state transitions. This design evaluates each candidate in the context of the evolved history while avoiding repeated history encoding. In addition, interaction histories may contain predictive patterns at different temporal scales, from slowly evolving preferences to rapid behavioral shifts. When modeled directly, dominant patterns at one scale may obscure weaker but predictive dynamics at another. We therefore introduce \textbf{Dual-Branch Representation Refinement (DRR)} to adaptively refine historical and candidate representations, providing CAME with more informative inputs for sequential encoding.

The main contributions of this work are summarized as follows:
\begin{itemize}

\item \textbf{A unified view of sequential dynamics.} 
We characterize sequential dynamics for future link prediction through two complementary signals: \emph{historical-sequence evolution} and \emph{candidate--history compatibility}. Through empirical analysis, we show that both signals contribute to strong performance across repeated- and unseen-edge prediction.

\item \textbf{An efficient candidate-aware Mamba-based framework.}
We propose \textbf{MambaTG}, which jointly captures these two signals through a Candidate-Aware Mamba Encoder and improves input representations through Dual-Branch Representation Refinement. By reusing the encoded historical state across candidates, MambaTG avoids repeated history encoding and enables efficient candidate evaluation.

\item \textbf{Extensive empirical validation.}
We evaluate MambaTG on 13 future link prediction datasets. MambaTG achieves strong performance across repeated- and unseen-edge prediction and attains the best average rank among all baselines. Efficiency and ablation studies further demonstrate its scalability and validate its principal design choices.
\end{itemize}

\section{Related Work}

\paragraph{Temporal Graph Learning Methods for Future Link Prediction.}
Temporal graph models use diverse architectures to encode interaction histories, including memory-based methods such as JODIE~\citep{jodie}, DyRep~\citep{DyRep}, and TGN~\citep{tgns}, attention-based methods such as TGAT~\citep{tgat}, TCL~\citep{tcl}, DyGFormer~\citep{dygformer}, and SimpleDyG~\citep{SimpleDyG}, and alternative encoders such as CAWN~\citep{cawn}, GraphMixer~\citep{graphmixer}, and FreeDyG~\citep{tian2024freedyg}. From the perspective of sequential dynamics, these methods differ in how they capture \emph{historical-sequence evolution} and \emph{candidate--history compatibility}. Many methods encode the source history and candidate largely independently and combine their representations only at the prediction stage~\citep{jodie,DyRep,tgns,tgat,graphmixer,SimpleDyG}, limiting direct candidate conditioning of the source history. Other approaches incorporate information from both endpoints through temporal motifs, cross-subgraph interactions, neighbor co-occurrence, or pairwise statistics~\citep{cawn,tcl,dygformer,tian2024freedyg}, but primarily characterize relationships between the source and destination histories rather than how a candidate relates to the source's evolving history. CRAFT~\citep{craft} directly matches each candidate against the source node's individual historical interactions, but does not explicitly model dependencies arising from the history's sequential evolution before candidate matching. In contrast, MambaTG jointly models \emph{historical-sequence evolution} and \emph{candidate--history compatibility} by first encoding the ordered source history to model its evolution and then adapting the resulting state to each candidate.

\paragraph{Mamba-Based Models for Continuous-Time Dynamic Graphs.}
Mamba~\citep{gu2024mambalineartimesequencemodeling} is a selective state-space model that efficiently models long sequences through input-dependent state transitions with linear sequence-length complexity. Recent studies have adapted Mamba to continuous-time dynamic graphs. DyGMamba~\citep{ding2025dygmambaefficientlymodelinglongterm} uses node- and time-level Mamba to encode long histories and identify temporally relevant interactions, while DyG-Mamba~\citep{li2025dyg-mambacontinuousstatespace} incorporates irregular inter-event times into state transitions for elapsed-time-aware forgetting and selective retention. These methods primarily focus on efficient long-history encoding and irregular temporal dynamics. In contrast, MambaTG uses candidate-aware continuation from the evolving source history to jointly model \emph{historical-sequence evolution} and \emph{candidate--history compatibility}, targeting effective prediction across both repeated and unseen edges.

\section{Preliminaries}

\paragraph{Temporal Graph and Future Link Prediction.}
A temporal graph is represented as
\(\mathcal{G}=(\mathcal{V},\mathcal{E})\), where \(\mathcal{V}\) is the node set and
\(\mathcal{E}=((u_i,v_i,t_i))_{i=1}^{M}\) is a chronologically ordered stream of \(M\) interactions. Here, \(u_i\in\mathcal{V}\) and \(v_i\in\mathcal{V}\) denote the source and destination nodes of the \(i\)-th interaction, respectively, and \(t_i\) denotes its timestamp, with \(t_1\leq t_2\leq\cdots\leq t_M\). At a query time \(t_q\), the observed history is
\(\mathcal{E}_{<t_q}=\{(u_i,v_i,t_i)\in\mathcal{E}\mid t_i<t_q\}\).
Given a source node \(u\) and a candidate set \(\mathcal{C}_{u}^{t_q}\), future link prediction assigns each candidate destination \(c_i\in\mathcal{C}_{u}^{t_q}\) a score conditioned on \(\mathcal{E}_{<t_q}\) and ranks the true destination above negative candidates. A candidate edge \((u,c_i)\) is \emph{repeated} if the pair appears in \(\mathcal{E}_{<t_q}\), and \emph{unseen} otherwise~\citep{yi2025tgbseqbenchmarkchallengingtemporal}.

\paragraph{Discrete Fourier Transform.}
Let \(\mathbf{X}=(\mathbf{x}_n)_{n=0}^{N-1}\in\mathbb{R}^{N\times d}\) be a sequence of \(d\)-dimensional representations. The discrete Fourier transform (DFT) and its inverse (IDFT), applied along the temporal dimension, are defined as
\begin{equation}
\widehat{\mathbf{x}}_m=\sum_{n=0}^{N-1}\mathbf{x}_n\exp\left(-\mathrm{i}\frac{2\pi mn}{N}\right),\qquad
\mathbf{x}_n=\frac{1}{N}\sum_{m=0}^{N-1}\widehat{\mathbf{x}}_m\exp\left(\mathrm{i}\frac{2\pi mn}{N}\right),
\end{equation}
where \(\widehat{\mathbf{x}}_m\in\mathbb{C}^{d}\) is the \(m\)-th frequency component and \(\mathrm{i}^{2}=-1\). For real-valued inputs, the real FFT (rFFT) and its inverse (irFFT) are commonly used~\citep{sorensen1987real}.

\paragraph{State Space Models.}
\label{sec:ssm}
State space models (SSMs) represent sequences through latent states that evolve over time. Given \(x(t)\in\mathbb{R}\), an SSM produces \(y(t)\in\mathbb{R}\) from a latent state \(h(t)\in\mathbb{R}^{N}\):
\begin{equation}
h'(t)=Ah(t)+Bx(t), \qquad y(t)=Ch(t).
\end{equation}
Here, \(A\in\mathbb{R}^{N\times N}\) is the state-transition matrix, \(B\in\mathbb{R}^{N\times 1}\) projects the input into the latent state, and \(C\in\mathbb{R}^{1\times N}\) maps the latent state to the output. To process a discrete sequence, the continuous system is discretized using a step size \(\Delta\):
\begin{equation}
h_t=\bar{A}h_{t-1}+\bar{B}x_t, \qquad y_t=Ch_t.
\end{equation}

Under zero-order-hold discretization, the discrete parameters are given by \(\bar{A}=\exp(\Delta A)\) and \(\bar{B}=(\Delta A)^{-1}\bigl(\exp(\Delta A)-I\bigr)\Delta B\). Unlike conventional SSMs with input-independent parameters, Mamba uses input-dependent \(B\), \(C\), and \(\Delta\) with a hardware-aware selective scan for efficient recurrence computation  \citep{gu2024mambalineartimesequencemodeling}.
\section{Method}
\begin{figure}[h]
    \centering
    \includegraphics[width=\linewidth]
    {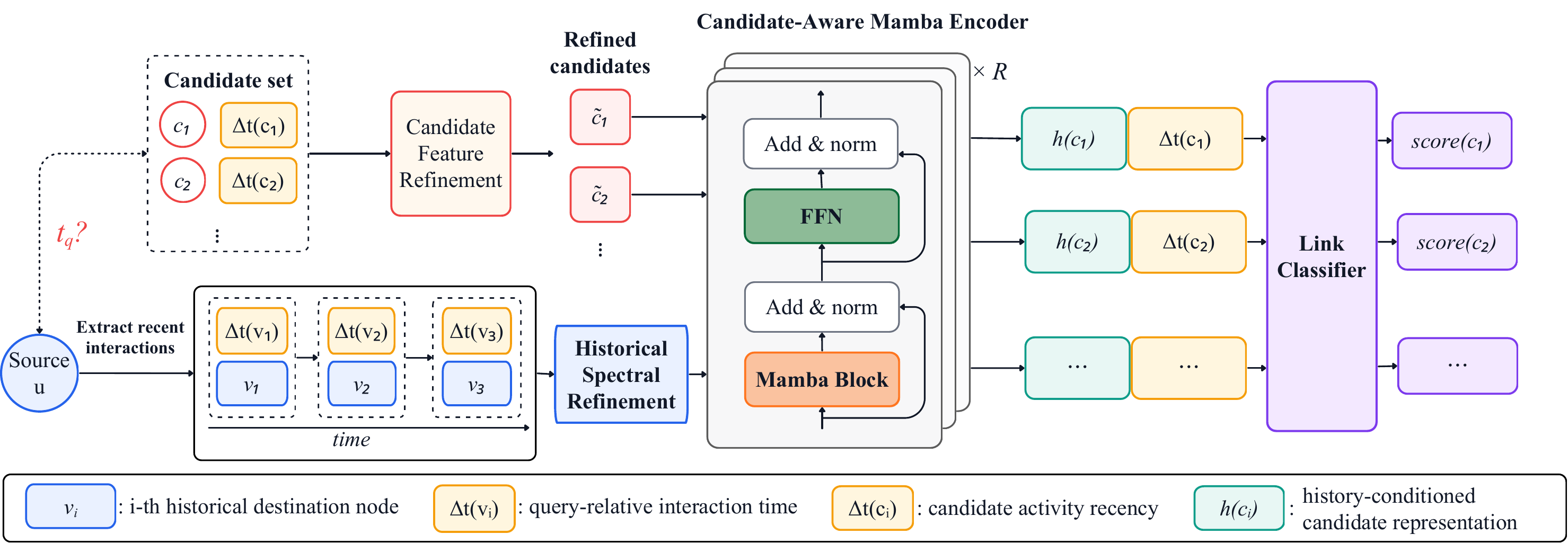}
    \caption{Framework of the proposed MambaTG model.}
    \label{fig:framework}
\end{figure}

The overall MambaTG framework is illustrated in Figure~\ref{fig:framework}. Given a source node \(u\), a query timestamp \(t_q\), and its chronologically ordered interaction history \(\mathcal{H}_{u}^{t_q}=((v_1,t_1),\ldots,(v_{\ell_u},t_{\ell_u}))\), where \(\ell_u\leq L_{\max}\) and \(t_1\leq\cdots\leq t_{\ell_u}<t_q\), MambaTG assigns a link-prediction score to each candidate \(c_i\in\mathcal{C}_{u}^{t_q}\). It first combines node embeddings with query-relative time encodings to construct historical and candidate representations. Dual-Branch Representation Refinement then enhances these representations through separate historical and candidate refinement branches, after which Candidate-Aware Mamba Encoder recurrently encodes the refined history and incorporates each candidate through an additional state update. Finally, a link classifier maps each history-conditioned candidate representation to a prediction score.

\subsection{Node and Time Encoding.}
\paragraph{Node embedding.}
Following \cite{craft}, we associate node \(v\in\mathcal{V}\) with a trainable identifier embedding \(\operatorname{Emb}(v)\in\mathbb{R}^{d_e}\). Given a query time \(t_q\), let \(\mathcal{C}_{u}^{t_q}\) denote the set of \(q=|\mathcal{C}_{u}^{t_q}|\) candidate destinations. We arrange the embeddings of the \(\ell_u\) historical neighbors and \(q\) candidate destinations row-wise as \(\mathbf{E}_H=[\operatorname{Emb}(v_i)]_{i=1}^{\ell_u}\in\mathbb{R}^{\ell_u\times d_e}\) and \(\mathbf{E}_D=[\operatorname{Emb}(c_j)]_{j=1}^{q}\in\mathbb{R}^{q\times d_e}\), respectively.

\paragraph{Time encoding.}\label{time_encoding}
To incorporate temporal information, we encode elapsed times relative to the query time \(t_q\). For each historical interaction occurring at time \(t_i\), its elapsed time is \(\Delta t_i^H=t_q-t_i\). For candidate \(c_j\), we instead use its activity recency \(\Delta t_{c_j}^D=t_q-t_{c_j}^{\mathrm{last}}\), where \(t_{c_j}^{\mathrm{last}}<t_q\) is its most recent interaction with any node in the global graph. For cold-start candidates with no prior interactions, we set \(\Delta t_{c_j}^D=0\). Following \cite{tgat}, each elapsed time is mapped to a \(d_T\)-dimensional representation using \(\boldsymbol{\phi}:\mathbb{R}_{\geq0}\rightarrow\mathbb{R}^{d_T}\). The resulting historical and candidate time encodings are \(\mathbf{T}_H=[\boldsymbol{\phi}(\Delta t_i^H)]_{i=1}^{\ell_u}\in\mathbb{R}^{\ell_u\times d_T}\) and \(\mathbf{T}_D=[\boldsymbol{\phi}(\Delta t_{c_j}^D)]_{j=1}^{q}\in\mathbb{R}^{q\times d_T}\), respectively.

\paragraph{Feature fusion.}
To jointly capture node identity and temporal context, we combine the node and time representations: \(\mathbf{Z}_H=\operatorname{Linear}([\mathbf{E}_H\Vert\mathbf{T}_H])\in\mathbb{R}^{\ell_u\times d}\) and \(\mathbf{Z}_D=\operatorname{Linear}([\mathbf{E}_D\Vert\mathbf{T}_D])\in\mathbb{R}^{q\times d}\), where \(\Vert\) denotes feature-wise concatenation and \(\operatorname{Linear}(\cdot)\) is a shared linear projection.

\subsection{Dual-Branch Representation Refinement (DRR).}
An interaction history may simultaneously contain slowly evolving trends, recurring behavior, and abrupt short-term changes. Without prior refinement, stronger patterns at one scale may obscure subtler yet predictive dynamics at other scales. We therefore introduce \textbf{Dual-Branch Representation Refinement}, comprising \textbf{Historical Spectral Refinement (HSR)} and \textbf{Candidate Feature Refinement (CFR)}. HSR exposes and adaptively refines frequency components associated with these patterns, while CFR recalibrates candidate features for compatibility with the refined history.

\paragraph{Historical Spectral Refinement.}
We transform $\mathbf{Z}_H$ into the frequency domain along its temporal dimension, yielding
$\widehat{\mathbf{Z}}_H=\operatorname{rFFT}_{L_{\max}}(\mathbf{Z}_H)\in\mathbb{C}^{(\lfloor L_{\max}/2\rfloor+1)\times d}$, where $L_{\max}\geq\ell_u$ is the transform length and zero-padding is applied when $\ell_u<L_{\max}$. Inspired by spectral filtering in FreeDyG~\citep{tian2024freedyg}, HSR modulates each frequency component using learnable complex-valued weights:

\begin{equation}
\widehat{\mathbf{Z}}_H^{\mathrm{HSR}}
=
\widehat{\mathbf{Z}}_H\odot\mathbf{W}_H,
\qquad
\mathbf{W}_H\in\mathbb{C}^{(\lfloor L_{\max}/2\rfloor+1)\times d}.
\end{equation}
Here, $\odot$ denotes element-wise multiplication. The complex-valued weights $\mathbf{W}_H$ enable the model to selectively adjust the magnitude and phase of temporal patterns across different frequencies. Applying the inverse transform with output length $L_{\max}$ yields
$\mathbf{Z}_H^{\mathrm{HSR}}
=
\left[\operatorname{irFFT}_{L_{\max}}
\left(\widehat{\mathbf{Z}}_H^{\mathrm{HSR}}\right)\right]_{1:\ell_u}
\in\mathbb{R}^{\ell_u\times d}$,
where the first $\ell_u$ positions corresponding to the original history are retained.

\paragraph{Candidate Feature Refinement.}
To promote alignment between candidate representations and the spectrally refined historical representations, CFR learnably rescales candidate features through a lightweight feature-wise transformation. Specifically, for each candidate representation \(\mathbf{z}_{c_i}=\mathbf{Z}_D[i]\in\mathbb{R}^{d}\), we compute
\(
\mathbf{z}_{c_i}^{\mathrm{CFR}}
=
\mathbf{z}_{c_i}\odot\mathbf{w}_D
\)
where \(\mathbf{w}_D\in\mathbb{R}^{d}\) is a learnable vector shared across candidates.

\subsection{Candidate-Aware Mamba Encoder (CAME).}
Modeling sequential dynamics requires capturing both \emph{historical-sequence evolution} and \emph{candidate--history compatibility}. Mamba~\citep{gu2024mambalineartimesequencemodeling} is well suited to this objective because its selective recurrence summarizes ordered sequences into recurrent states while its input-dependent state transitions allow the same historical state to respond differently to different candidates. CAME stacks \(R\) layers, each consisting of a Mamba block followed by a position-wise feed-forward network (FFN). Detailed formulations of these standard components are provided in Appendix~\ref{app:mamba_block}.

\paragraph{Historical-sequence encoding.}
CAME first encodes the source node's interaction history in chronological order to capture its sequential evolution. The refined historical representations \(\mathbf{Z}_H^{\mathrm{HSR}}\in\mathbb{R}^{\ell_u\times d}\) are padded to \(L_{\max}\) for batching. We initialize \(\mathbf{X}_H^{(0)}=\operatorname{Pad}\!\left(\mathbf{Z}_H^{\mathrm{HSR}},L_{\max}\right)\). At each layer \(r\), CAME processes the complete historical sequence once and retains its terminal recurrent cache:
\begin{equation}
\begin{gathered}
\left(\mathbf{M}_H^{(r)},\mathcal{K}_H^{(r)}\right)
=
\operatorname{MambaBlock}^{(r)}
\left(\mathbf{X}_H^{(r-1)}\right),
\qquad
\widetilde{\mathbf{X}}_H^{(r)}
=
\operatorname{LN}_1^{(r)}
\left(
\mathbf{X}_H^{(r-1)}+\mathbf{M}_H^{(r)}
\right),\\
\mathbf{X}_H^{(r)}
=
\operatorname{LN}_2^{(r)}
\left(
\widetilde{\mathbf{X}}_H^{(r)}
+
\operatorname{FFN}^{(r)}
\left(\widetilde{\mathbf{X}}_H^{(r)}\right)
\right),
\qquad r=1,\ldots,R.
\end{gathered}
\end{equation}

Here, \(\mathbf{M}_H^{(r)}\) denotes the Mamba output representations for the historical sequence, while \(\mathcal{K}_H^{(r)}\) contains the terminal SSM and causal-convolution states required to continue the \(r\)-th Mamba layer with one additional token. Through recurrent state propagation, historical representations incorporate information from earlier interactions, enabling CAME to capture \emph{historical-sequence evolution}.

\paragraph{Candidate continuation.}
For candidate \(c_i\), we initialize \(\mathbf{x}_{c_i}^{(0)}=\mathbf{z}_{c_i}^{\mathrm{CFR}}\). At each layer, the candidate is processed as one additional input following the historical sequence by continuing from the corresponding terminal history cache:
\begin{equation}
\begin{gathered}
\mathbf{m}_{c_i}^{(r)}
=
\operatorname{MambaStep}^{(r)}
\left(
\mathbf{x}_{c_i}^{(r-1)};
\mathcal{K}_H^{(r)}
\right),
\qquad
\widetilde{\mathbf{x}}_{c_i}^{(r)}
=
\operatorname{LN}_1^{(r)}
\left(
\mathbf{x}_{c_i}^{(r-1)}
+
\mathbf{m}_{c_i}^{(r)}
\right),\\
\mathbf{x}_{c_i}^{(r)}
=
\operatorname{LN}_2^{(r)}
\left(
\widetilde{\mathbf{x}}_{c_i}^{(r)}
+
\operatorname{FFN}^{(r)}
\left(
\widetilde{\mathbf{x}}_{c_i}^{(r)}
\right)
\right),
\qquad r=1,\ldots,R.
\end{gathered}
\end{equation}

Here, \(\operatorname{MambaStep}^{(r)}\) shares its parameters with \(\operatorname{MambaBlock}^{(r)}\) and continues the recurrence for one additional input using \(\mathcal{K}_H^{(r)}\). Since the selective state-space parameters depend on the candidate input \(\mathbf{x}_{c_i}^{(r-1)}\), each candidate induces a different state update from the shared historical cache. The resulting representation is therefore conditioned jointly on the candidate and the evolved historical state, capturing \emph{candidate--history compatibility}. By reusing the terminal history cache across candidates, MambaTG encodes the history only once, while each candidate requires only one additional recurrent update per layer, avoiding redundant history encoding. After \(R\) layers, the history-conditioned candidate representation is \(\mathbf{h}_{u,c_i}^{t_q}=\mathbf{x}_{c_i}^{(R)}\in\mathbb{R}^{d}\).

\subsection{Link Classifier.}
Following \cite{craft}, we project the candidate's activity recency, \(\Delta t_{c_i}^{D}=t_q-t_{c_i}^{\mathrm{last}}\), and concatenate it with the candidate representation \(\mathbf{h}_{u,c_i}^{t_q}\). Prediction score is computed as:
\begin{equation}
\hat{y}_{u,c_i}^{t_q}
=
\operatorname{MLP}\!\left(
\left[
\mathbf{h}_{u,c_i}^{t_q}
\Vert
\operatorname{Linear}_{\Delta}\!\left(\Delta t_{c_i}^{D}\right)
\right]
\right),
\end{equation}
where \(\Vert\) denotes feature-wise concatenation. For repeated-edge-dominant datasets, we use \textbf{MambaTG-R}, which additionally projects and concatenates the historical repetition frequency \(n_{u,c_i}^{t_q}\), defined as the number of interactions between \(u\) and \(c_i\) before \(t_q\). Both variants are optimized using Bayesian Personalized Ranking (BPR)~\citep{rendle2012bprbayesianpersonalizedranking}, which encourages observed destinations to receive higher scores than sampled negatives.

\subsection{Time Complexity Analysis}

\begin{table}[h]
\centering
\caption{Candidate-scoring time complexity. Entries omit \(O(\cdot)\) and fixed architectural constants.}
\label{tab:time_complexity}
\begin{tabular}{lcc}
\toprule
Method & Historical-Neighbor Extraction & Edge Likelihood Computation \\
\midrule
TGN
& \(B(1+q)\log\bar{d}\)
& \(B(1+q)\ell d^2\) \\
TGAT
& \(B(1+q)(1+\ell)\log\bar{d}\)
& \(B(1+q)\ell^2d^2\) \\
GraphMixer
& \(B(1+q)\log\bar{d}\)
& \(B(1+q)\ell d^2\) \\
DyGFormer
& \(B(1+q)\log\bar{d}\)
& \(B(1+q)(\ell d^2+\ell^2d)\) \\
DyG-Mamba
& \(B(1+q)\log\bar{d}\)
& \(Bq\ell d^2\) \\
DyGMamba
& \(B(1+q)\log\bar{d}\)
& \(Bq\ell d^2\) \\
CRAFT
& \(B\log\bar{d}\)
& \(B(\ell+q)d^2+Bq\ell d\) \\
\textbf{MambaTG}
& \(B\log\bar{d}\)
& \(B(\ell+q)d^2+B\ell d\log \ell\) \\
\bottomrule
\end{tabular}
\end{table}

We analyze the time complexity of future link prediction for a batch of source nodes and their candidate destinations. Let \(B\) denote the number of source queries, \(q\) the number of candidates per source, \(\ell\) the number of historical interactions considered per node, \(d\) the feature dimension, and \(\bar{d}\) the average node degree. Table~\ref{tab:time_complexity} summarizes the resulting complexities; detailed derivations and assumptions are provided in Appendix~\ref{app:complexity_derivations}.

MambaTG retrieves only the source history, requiring \(O(B\log\bar{d})\) time for temporal-index lookup. For edge likelihood computation, DRR refines each history in \(O(B\ell d\log\ell)\) time, while CAME encodes it once and reuses its layer-wise terminal caches across candidates, costing \(O(B(\ell+q)d^2)\). The total complexity is \(O\!\left(B(\ell+q)d^2+B\ell d\log\ell\right)\). Unlike methods that encode both endpoints, MambaTG uses only the source history, avoiding candidate-dependent neighborhood retrieval. The separation of history encoding from candidate evaluation further avoids repeated processing across candidates and yields more favorable asymptotic scaling than CRAFT when \(q>\log\ell\), as in our evaluation, with \(q=101\) and \(\ell\) up to 1000. Together, they enable scalable prediction over long histories and large candidate sets in applications such as recommendation and social-link prediction.

\section{Experiments}
\subsection{Experimental Setup}
\paragraph{Datasets.}
We evaluate MambaTG on 13 future link prediction datasets, including seven datasets from TGB-Seq~\citep{yi2025tgbseqbenchmarkchallengingtemporal} and six commonly used temporal graph datasets. Following \citet{craft}, we categorize them as repeated-edge-dominant or unseen-edge-dominant according to the relative prevalence of repeated and unseen edges. Dataset details are provided in Appendix~\ref{app:datasets}.

\paragraph{Baselines.}
We compare MambaTG with twelve representative temporal graph baselines: EdgeBank~\citep{poursafaei2022towards}, CRAFT~\citep{craft}, JODIE~\citep{jodie}, DyRep~\citep{DyRep}, TGAT~\citep{tgat}, TGN~\citep{tgns}, CAWN~\citep{cawn}, GraphMixer~\citep{graphmixer}, DyGFormer~\citep{dygformer}, FreeDyG~\citep{tian2024freedyg}, DyG-Mamba~\citep{li2025dyg-mambacontinuousstatespace}, and DyGMamba~\citep{ding2025dygmambaefficientlymodelinglongterm}. Following \citet{craft}, we additionally include SGNN-HN~\citep{pan2020star}, a session-based recommendation method, as a sequential baseline. Together, these methods span non-parametric, memory-based, attention-based, walk-based, MLP-based, frequency-based, Mamba-based, and sequential recommendation architectures. Baseline details are provided in Appendix~\ref{app:baseline_details}.

\begin{table}[h]
\centering
\caption{Future link prediction performance on
(a) unseen-edge-dominant and (b) repeated-edge-dominant datasets.
Results are reported as MRR (\%), averaged over three random seeds.
The \best{best}, \second{second-best}, and \third{third-best} results
are highlighted. Avg.Rk. denotes the average rank over datasets with
reported results (lower is better); parentheses indicate coverage in (a).}

\label{tab:main_results}

\begingroup
\scriptsize
\renewcommand{\arraystretch}{0.92}

\textbf{(a) Unseen-edge-dominant datasets}

\setlength{\tabcolsep}{1.1pt}
\begin{tabular*}{\textwidth}{
  @{\extracolsep{\fill}}lcccccccc@{}
}
\toprule
Model & ML-20M & Taobao & Yelp & GoogleLocal & Flickr & YouTube
& WikiLink & Avg.Rk. \\
\midrule
JODIE
& $21.16{\pm}0.73$
& $48.36{\pm}2.18$
& \third{$69.88{\pm}0.31$}
& $41.86{\pm}1.49$
& $46.21{\pm}0.83$
& $41.67{\pm}2.86$
& $57.94{\pm}1.33$
& $6.43\,(7/7)$ \\

DyRep
& $19.00{\pm}1.69$
& $40.03{\pm}2.40$
& $57.69{\pm}1.05$
& $37.73{\pm}1.34$
& $38.04{\pm}4.19$
& $35.12{\pm}4.13$
& $42.63{\pm}1.33$
& $8.29\,(7/7)$ \\

TGAT
& $10.47{\pm}0.20$
& --
& --
& $19.78{\pm}0.24$
& $23.53{\pm}3.35$
& $43.56{\pm}2.53$
& --
& $10.75\,(4/7)$ \\

TGN
& $23.99{\pm}0.20$
& $60.28{\pm}0.54$
& $69.79{\pm}0.24$
& $54.13{\pm}1.97$
& $46.03{\pm}6.78$
& $55.16{\pm}5.89$
& $62.94{\pm}2.16$
& $5.29\,(7/7)$ \\

CAWN
& $12.31{\pm}0.02$
& --
& $25.71{\pm}0.09$
& $18.26{\pm}0.02$
& $48.69{\pm}6.08$
& $47.55{\pm}1.08$
& --
& $9.00\,(5/7)$ \\

GraphMixer
& $21.97{\pm}0.17$
& $31.54{\pm}0.02$
& $33.96{\pm}0.19$
& $21.31{\pm}0.14$
& $45.01{\pm}0.08$
& $58.87{\pm}0.12$
& $48.57{\pm}0.02$
& $6.86\,(7/7)$ \\

DyGFormer
& --
& --
& $21.68{\pm}0.20$
& $18.39{\pm}0.02$
& $49.58{\pm}2.87$
& $46.08{\pm}3.44$
& --
& $9.00\,(4/7)$ \\

DyGMamba
& $5.86{\pm}0.21$
& --
& --
& $17.64{\pm}0.06$
& $52.07{\pm}0.43$
& $54.09{\pm}1.68$
& --
& $9.25\,(4/7)$ \\

DyG-Mamba
& --
& --
& --
& $17.34{\pm}0.36$
& $55.86{\pm}0.14$
& $56.71{\pm}0.23$
& --
& $8.00\,(3/7)$ \\

FreeDyG
& $20.67{\pm}0.70$
& --
& --
& $20.49{\pm}0.13$
& $56.22{\pm}0.33$
& \best{$60.35{\pm}0.21$}
& --
& $5.00\,(4/7)$ \\

SGNN-HN
& \third{$33.12{\pm}0.01$}
& \third{$68.58{\pm}0.21$}
& $69.34{\pm}0.44$
& \second{$62.88{\pm}0.51$}
& \third{$60.15{\pm}0.20$}
& \third{$59.64{\pm}0.22$}
& \third{$69.37{\pm}0.39$}
& \third{$3.14\,(7/7)$} \\

CRAFT
& \second{$35.91{\pm}0.65$}
& \second{$70.68{\pm}0.42$}
& \second{$72.69{\pm}0.61$}
& \third{$62.35{\pm}0.46$}
& \second{$62.34{\pm}0.84$}
& $58.92{\pm}0.16$
& \best{$75.48{\pm}0.84$}
& \second{$2.29\,(7/7)$} \\

\textbf{MambaTG}
& \best{$37.98{\pm}0.11$}
& \best{$73.06{\pm}0.21$}
& \best{$75.85{\pm}0.17$}
& \best{$64.62{\pm}0.45$}
& \best{$63.94{\pm}0.72$}
& \second{$59.81{\pm}0.74$}
& \second{$75.38{\pm}0.13$}
& \best{$1.29\,(7/7)$} \\
\bottomrule
\end{tabular*}

\medskip
\textbf{(b) Repeated-edge-dominant datasets}

\setlength{\tabcolsep}{1.8pt}
\begin{tabular*}{\textwidth}{
  @{\extracolsep{\fill}}lccccccc@{}
}
\toprule
Model & Wikipedia & Reddit & MOOC & LastFM & UCI & Flights & Avg.Rk. \\
\midrule
EdgeBank
& $77.94{\pm}0.00$
& $77.88{\pm}0.00$
& $9.76{\pm}0.00$
& $19.07{\pm}0.00$
& $39.43{\pm}0.00$
& $54.46{\pm}0.00$
& $11.50$ \\

JODIE
& $76.48{\pm}1.72$
& $77.16{\pm}1.27$
& $19.91{\pm}3.06$
& $18.49{\pm}2.87$
& $51.43{\pm}1.74$
& $20.37{\pm}4.18$
& $11.83$ \\

DyRep
& $67.42{\pm}4.21$
& $72.33{\pm}2.14$
& $17.71{\pm}1.27$
& $17.76{\pm}6.56$
& $14.00{\pm}2.71$
& $15.62{\pm}0.47$
& $13.67$ \\

TGAT
& $72.72{\pm}1.50$
& $78.03{\pm}0.25$
& $31.55{\pm}4.14$
& $24.10{\pm}1.26$
& $31.50{\pm}0.48$
& $53.90{\pm}1.07$
& $11.17$ \\

TGN
& $82.22{\pm}0.39$
& $79.25{\pm}0.24$
& $39.03{\pm}4.53$
& $25.59{\pm}4.43$
& $45.29{\pm}6.43$
& $64.24{\pm}2.11$
& $9.25$ \\

CAWN
& $86.07{\pm}0.08$
& $87.66{\pm}0.04$
& $27.02{\pm}0.57$
& $35.10{\pm}0.34$
& $66.96{\pm}0.48$
& $80.06{\pm}2.10$
& $8.00$ \\

GraphMixer
& $73.57{\pm}1.48$
& $70.87{\pm}0.38$
& $29.02{\pm}0.28$
& $26.76{\pm}1.42$
& $59.58{\pm}1.00$
& $42.26{\pm}0.49$
& $10.83$ \\

DyGFormer
& $88.69{\pm}0.04$
& $88.70{\pm}0.05$
& $42.20{\pm}1.31$
& $46.50{\pm}0.51$
& $76.61{\pm}0.26$
& \third{$84.13{\pm}3.46$}
& $4.83$ \\

DyGMamba
& \second{$89.10{\pm}0.17$}
& $89.00{\pm}0.03$
& $47.17{\pm}1.43$
& $46.39{\pm}0.49$
& \third{$76.98{\pm}0.46$}
& \best{$85.02{\pm}0.28$}
& \second{$3.50$} \\

DyG-Mamba
& $88.50{\pm}0.21$
& $88.77{\pm}0.14$
& $47.33{\pm}0.23$
& \third{$53.42{\pm}0.19$}
& \second{$78.25{\pm}0.60$}
& $81.47{\pm}0.04$
& \third{$4.00$} \\

FreeDyG
& \best{$90.18{\pm}0.12$}
& \third{$89.16{\pm}0.05$}
& $44.51{\pm}0.37$
& $49.19{\pm}0.55$
& $74.55{\pm}1.12$
& $81.10{\pm}0.06$
& $4.33$ \\

SGNN-HN
& $82.22{\pm}0.35$
& $87.42{\pm}0.04$
& \third{$57.25{\pm}0.17$}
& $40.72{\pm}0.26$
& $57.23{\pm}0.39$
& $80.80{\pm}0.14$
& $7.08$ \\

CRAFT-R
& $88.25{\pm}0.26$
& \second{$89.33{\pm}0.11$}
& \second{$62.32{\pm}0.39$}
& \second{$55.21{\pm}0.23$}
& $75.11{\pm}0.09$
& $83.16{\pm}0.16$
& \second{$3.50$} \\

\textbf{MambaTG-R}
& \third{$88.99{\pm}0.02$}
& \best{$89.71{\pm}0.10$}
& \best{$64.77{\pm}0.22$}
& \best{$56.98{\pm}0.24$}
& \best{$79.31{\pm}0.26$}
& \second{$84.51{\pm}0.03$}
& \best{$1.50$} \\
\bottomrule
\end{tabular*}

\endgroup
\end{table}

\paragraph{Evaluation Settings and Metrics.}
Following TGB-Seq~\citep{yi2025tgbseqbenchmarkchallengingtemporal}, we use Mean Reciprocal Rank (MRR)~\citep{craswell2009mrr} as the evaluation metric and adopt official negative samples when available. Otherwise, we randomly sample 100 negative destinations per positive interaction. All datasets are split chronologically into 70\%/15\%/15\% for training, validation, and testing, respectively. We report the average performance over three random seeds. Further evaluation and implementation details are provided in Appendix~\ref{app:implementation_details}.

\subsection{Predictive Effectiveness and Computational Efficiency}
\label{sec:effectiveness}
\label{sec:efficiency}

\paragraph{Overall Predictive Performance.}
Table~\ref{tab:main_results} reports MRR across all datasets. We use MambaTG for unseen-edge-dominant datasets and MambaTG-R for repeated-edge-dominant datasets, with the latter additionally incorporating source--candidate repetition frequency. EdgeBank, FreeDyG, DyG-Mamba, and DyGMamba results are from our experiments, while the remaining baseline results are taken from prior studies~\citep{craft,yi2025tgbseqbenchmarkchallengingtemporal}. Missing entries indicate experiments exceeding the specified runtime or GPU memory budget. Further details are provided in Appendix~\ref{app:implementation_details}.

\begin{wrapfigure}{r}{0.48\textwidth}
    \vspace*{-10pt}
    \centering
    \includegraphics[width=0.75\linewidth, keepaspectratio]{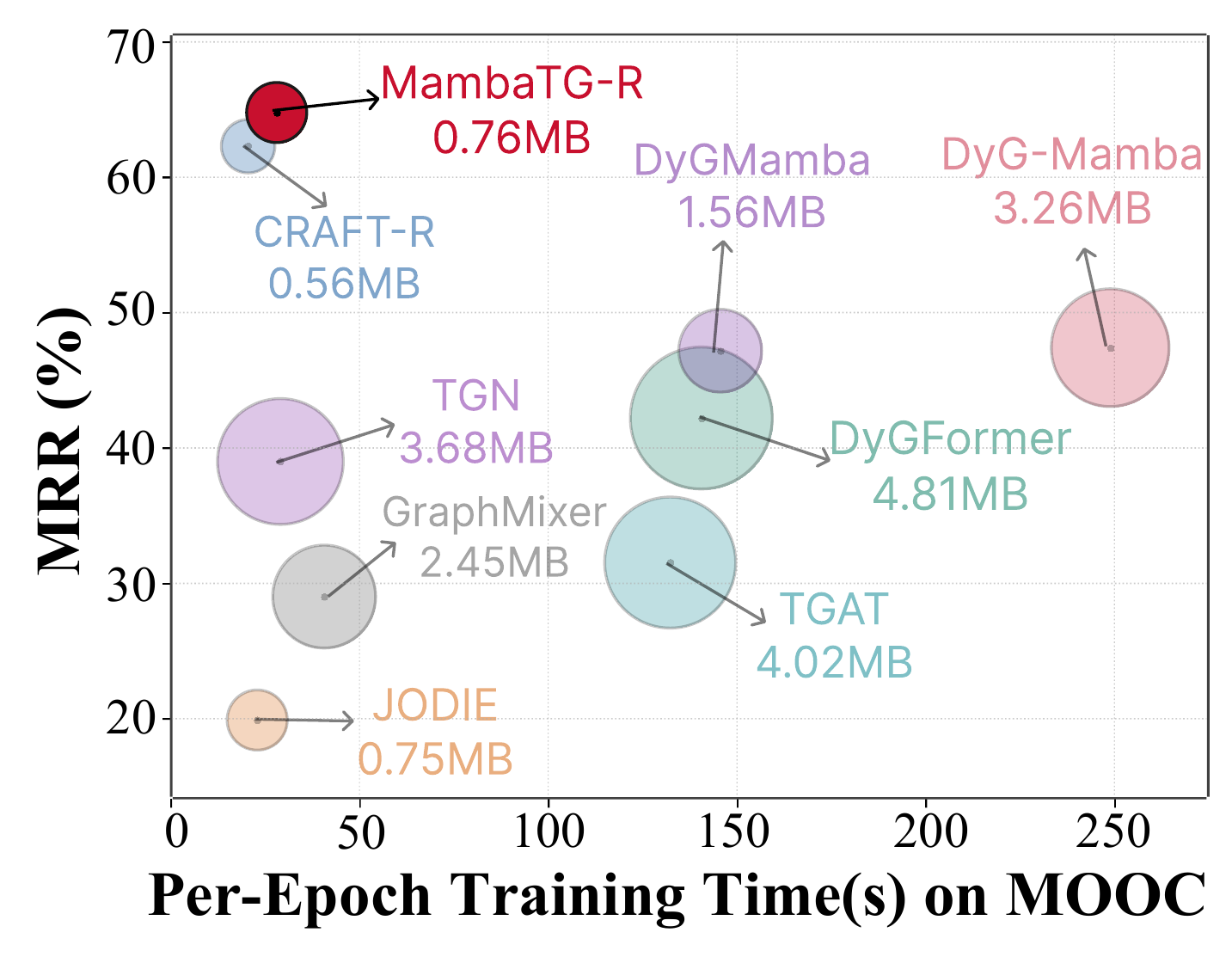}
    \caption{Performance versus efficiency on MOOC. Bubble size indicates model size.}
    \label{fig:pareto_analysis}
    \vspace{-20pt}
\end{wrapfigure}

MambaTG and MambaTG-R achieve the best performance on 9 of the 13 datasets, with strong performance across the remaining datasets, and attain the best average rank in both regimes: \(1.29\) on unseen-edge-dominant datasets and \(1.50\) on repeated-edge-dominant datasets. We attribute these gains primarily to two design choices: (i) CAME jointly captures \emph{historical-sequence evolution} and \emph{candidate--history compatibility}, supporting prediction across repeated and unseen edges, as validated in Section~\ref{sec:signal_ablation} and Appendix~\ref{app:came_impact}; and (ii) DRR refines historical patterns associated with different temporal scales while promoting compatibility between the refined history and candidate representations, as examined in Section~\ref{sec:component_ablation} and Appendix~\ref{app:drr_impact}. Moreover, layer-wise caching encodes each source history once and reuses its terminal states across candidates, enabling MambaTG to scale to large datasets on which several competing methods exceed the computational budget, as analyzed in Figure ~\ref{fig:pareto_analysis} and Appendix~\ref{app:additional_efficiency_results}.

\paragraph{Long-Sequence Effectiveness.}
\label{sec:long_sequence}

\begin{wrapfigure}{r}{0.51\textwidth}
    \vspace*{-14pt}
    \centering
    \includegraphics[
        width=0.75\linewidth,
        keepaspectratio
    ]{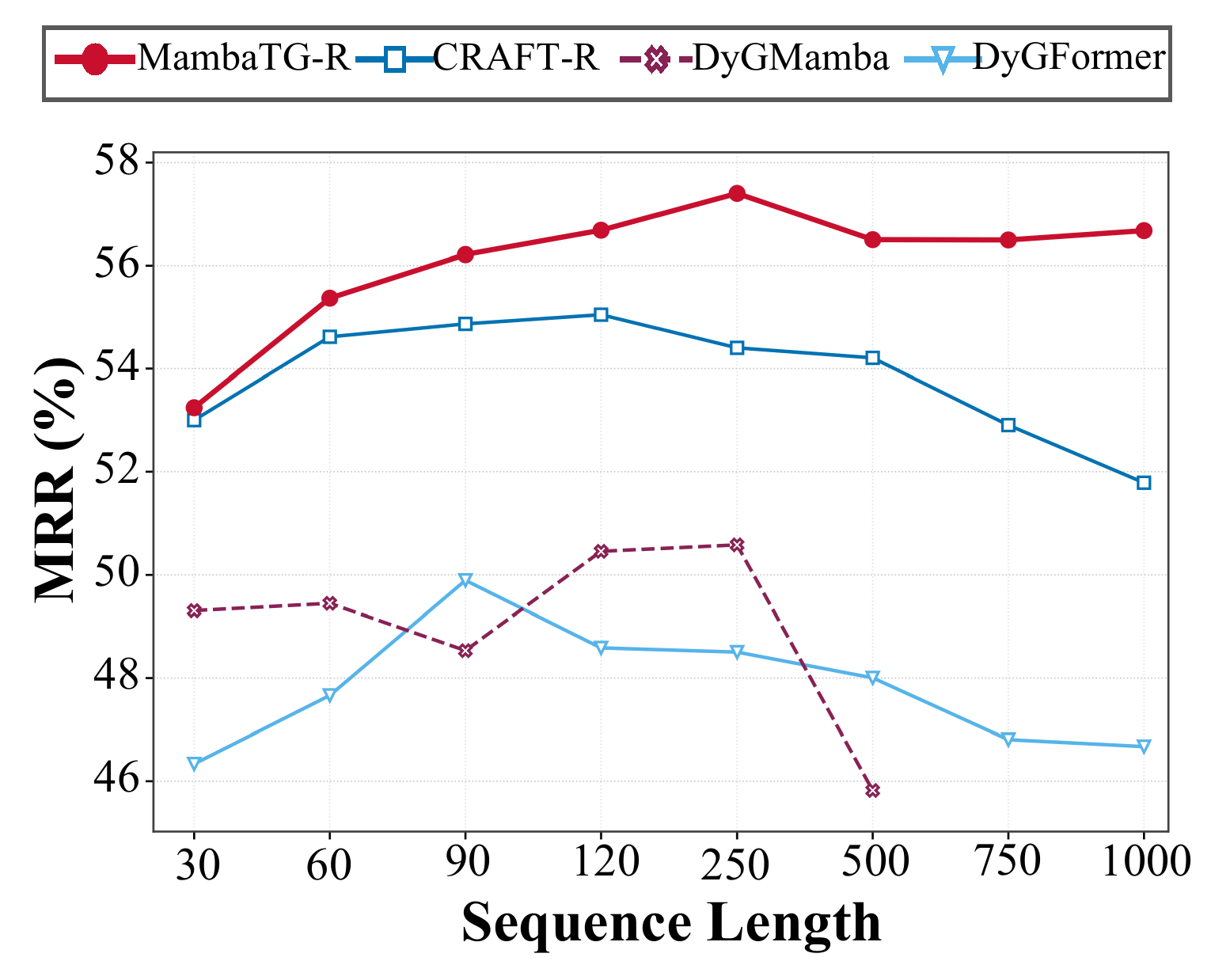}
    \caption{Long-sequence effectiveness on LastFM.}
    \label{fig:lastfm_long_sequence}
    \vspace{-20pt}
\end{wrapfigure}

To assess MambaTG's ability to exploit long interaction histories, we evaluate it on LastFM, which benefits from longer historical sequences. We vary the maximum source-history length \(\ell\) and compare MambaTG-R with other representative baselines.\footnote{DyGMamba results at \(\ell=750\) and \(\ell=1000\) are omitted because training became numerically unstable and produced NaN prediction scores.} As shown in Figure~\ref{fig:lastfm_long_sequence}, MambaTG-R maintains strong performance as \(\ell\) increases, while CRAFT-R and DyGFormer degrade with longer histories. These results demonstrate MambaTG's robustness to increasing history lengths and its ability to leverage extended temporal context effectively.
\finishwrap
\vspace{-1.5\baselineskip}

\paragraph{Performance--Efficiency Trade-off.}

Figure~\ref{fig:pareto_analysis} illustrates the trade-off between predictive performance and computational efficiency on MOOC. MambaTG-R consistently achieves the highest test MRR while remaining competitive in parameter size and per-epoch training time. In particular, MambaTG-R achieves up to \(8.9\times\) faster per-epoch training and a \(6.3\times\) smaller parameter footprint than competing methods. Overall, MambaTG-R offers the most favorable trade-off between predictive performance and computational efficiency across the evaluated datasets. Results on UCI and LastFM in Appendix~\ref{app:performance_efficiency_trade_off}.

\subsection{Validating Historical-Sequence Evolution and Candidate--History Compatibility}
\label{sec:signal_ablation}

\begin{wraptable}{r}{0.45\textwidth}
\vspace{-1.1\intextsep}
\centering
\caption{Macro-average MRR change (pp) relative to full MambaTG under signal ablations.}
\label{tab:signal_ablation_drop}
\small
\setlength{\tabcolsep}{4pt}
\renewcommand{\arraystretch}{0.95}
\begin{tabular}{lcc}
\toprule
Variant & Repeated & Unseen \\
\midrule
ShuffledTrain     & \dropval{8.23}  & \dropval{1.56} \\
ShuffledInference & \dropval{27.21} & \dropval{3.53} \\
ShuffledBoth      & \dropval{4.22}  & \dropval{1.78} \\
LateFusion        & \dropval{2.88}  & \dropval{3.98} \\
\bottomrule
\end{tabular}
\vspace{-1.2\intextsep}
\end{wraptable}

Our hypothesis is that MambaTG's effectiveness arises from jointly modeling \emph{historical-sequence evolution} and \emph{candidate--history compatibility}. To validate this hypothesis, we examine whether performance decreases when (i) \emph{the chronological order of the source node's interaction history is disrupted} and (ii) \emph{candidate-conditioned state updating is replaced with separate history--candidate fusion}.

For (i), following \citet{kim2025lost}, we adopt \emph{ShuffledTrain} and \emph{ShuffledInference}, which permute the source node's interaction history only during training or evaluation, respectively. We additionally introduce \emph{ShuffledBoth}, which shuffles histories during both training and evaluation. Table~\ref{tab:signal_ablation_drop} shows lower macro-average MRR for all three variants in both dataset groups. In particular, the degradation under \emph{ShuffledBoth}, which preserves neighbor identities while disrupting chronological order, shows that chronological order provides predictive information beyond neighbor identity. Degradation under \emph{ShuffledInference} indicates that MambaTG's predictions depend on the chronological ordering of historical interactions. Together, these results show that MambaTG leverages \emph{historical-sequence evolution} rather than treating the history as an unordered set.

For (ii), we introduce \emph{LateFusion}, which retains the original history encoder but replaces the candidate-conditioned Mamba continuation with MLP-based fusion of the separately encoded history and candidate representations, a common prediction design in T-GNNs. The resulting degradation indicates that late-stage fusion does not fully recover the benefit of candidate continuation from the evolved historical state. The larger degradation on GoogleLocal and YouTube further suggests that \emph{candidate--history compatibility} is particularly important for unseen-edge prediction. Overall, these findings support our hypothesis that jointly modeling \emph{historical-sequence evolution} and \emph{candidate--history compatibility} yields MambaTG's strongest performance. Detailed settings and per-dataset results are provided in Appendix~\ref{app:signal_ablation_details}.

\subsection{Ablation Study}
\label{sec:component_ablation}

\begin{wraptable}{r}{0.5\linewidth}
\vspace{-1.1\intextsep}
\centering
\caption{Macro-average MRR change (pp) under component ablations. Repetition frequency applies only to repeated-edge-dominant datasets.}
\label{tab:component_ablation}
\footnotesize
\setlength{\tabcolsep}{3pt}
\begin{tabular}{@{}lcc@{}}
\toprule
\textbf{Removed component} & \textbf{Repeated} & \textbf{Unseen} \\
\midrule
HSR                          & \dropval{0.21} & \dropval{1.74} \\
CFR                          & \dropval{0.57} & \dropval{1.28} \\
Query-relative time encoding & \dropval{2.87} & \dropval{1.77} \\
Candidate activity recency   & \dropval{2.35} & \dropval{1.03} \\
Historical repetition freq.  & \dropval{5.67} & N/A            \\
\bottomrule
\end{tabular}
\vspace{-15pt}
\end{wraptable}

We ablate five components of MambaTG: (1) Historical Spectral Refinement (\textbf{HSR}) and (2) Candidate Feature Refinement (\textbf{CFR}) within DRR, (3) \textbf{Query-relative time encoding}, (4) \textbf{Candidate activity recency}, and (5) \textbf{Historical repetition frequency}. Table~\ref{tab:component_ablation} reports the macro-average MRR (\%) reduction across two repeated-edge-dominant (UCI, MOOC) and two unseen-edge-dominant (GoogleLocal, Flickr) datasets. Variant definitions and per-dataset results are provided in Appendix~\ref{app:component_ablation}.

Overall, all components contribute to MambaTG, but their importance differs across prediction regimes. Query-relative time encoding, candidate activity recency, and historical repetition frequency contribute more strongly to repeated-edge prediction, where the timing and frequency of previous source--candidate interactions provide direct evidence of recurrence. In contrast, HSR and CFR show more significant contributions to unseen-edge prediction, where prediction relies more heavily on patterns extracted from the source history and informative candidate representations.
\section{Conclusion}
In this paper, we identified \emph{historical-sequence evolution} and \emph{candidate--history compatibility} as complementary sources of predictive information beyond direct pairwise recurrence. We proposed MambaTG, a candidate-aware Mamba-based framework that jointly models these signals to capture how ordered interaction histories evolve and relate to each candidate. Across 13 temporal graph datasets, MambaTG achieves strong performance in repeated-edge-dominant and unseen-edge-dominant settings while efficiently handling long histories and large candidate sets.

\subsection*{AI use statement}

We used generative AI tools to assist with code implementation, provide writing feedback, identify relevant literature, and provide feedback on experimental methodology. We reviewed the AI-assisted work and take responsibility for the final content.

\bibliography{iclr2027_conference}

@inproceedings{poursafaei2022towards,
	title = {Towards {Better} {Evaluation} for {Dynamic} {Link} {Prediction}},
	url = {https://openreview.net/forum?id=1GVpwr2Tfdg},
	language = {en},
	urldate = {2026-09-17},
	author = {Poursafaei, Farimah and Huang, Andy and Pelrine, Kellin and Rabbany, Reihaneh},
	month = jun,
	year = {2022},
}

@article{DBLP:journals/corr/abs-1905-11485,
	title = {Representation {Learning} for {Dynamic} {Graphs}: {A} {Survey}},
	volume = {21},
	issn = {1533-7928},
	shorttitle = {Representation {Learning} for {Dynamic} {Graphs}},
	url = {http://jmlr.org/papers/v21/19-447.html},
	number = {70},
	urldate = {2026-09-17},
	journal = {Journal of Machine Learning Research},
	author = {Kazemi, Seyed Mehran and Goel, Rishab and Jain, Kshitij and Kobyzev, Ivan and Sethi, Akshay and Forsyth, Peter and Poupart, Pascal},
	year = {2020},
	pages = {1--73},
}

@inproceedings{tgat,
	title = {Inductive representation learning on temporal graphs},
	url = {https://openreview.net/forum?id=rJeW1yHYwH},
	language = {en},
	urldate = {2026-09-02},
	author = {Xu, Da and Ruan, Chuanwei and Korpeoglu, Evren and Kumar, Sushant and Achan, Kannan},
	year = {2020},
}

@inproceedings{cawn,
	title = {Inductive {Representation} {Learning} in {Temporal} {Networks} via {Causal} {Anonymous} {Walks}},
	url = {https://openreview.net/forum?id=KYPz4YsCPj},
	language = {en},
	urldate = {2026-09-02},
	author = {Wang, Yanbang and Chang, Yen-Yu and Liu, Yunyu and Leskovec, Jure and Li, Pan},
	year = {2021},
}

@inproceedings{DyRep,
	title = {{DyRep}: {Learning} {Representations} over {Dynamic} {Graphs}},
	shorttitle = {{DyRep}},
	url = {https://openreview.net/forum?id=HyePrhR5KX},
	language = {en},
	urldate = {2026-09-02},
	author = {Trivedi, Rakshit and Farajtabar, Mehrdad and Biswal, Prasenjeet and Zha, Hongyuan},
	year = {2019},
}

@inproceedings{tian2024freedyg,
	title = {{FreeDyG}: {Frequency} {Enhanced} {Continuous}-{Time} {Dynamic} {Graph} {Model} for {Link} {Prediction}},
	shorttitle = {{FreeDyG}},
	url = {https://openreview.net/forum?id=82Mc5ilInM},
	language = {en},
	urldate = {2026-09-02},
	author = {Tian, Yuxing and Qi, Yiyan and Guo, Fan},
	year = {2024},
}

@inproceedings{gu2024mambalineartimesequencemodeling,
	title = {Mamba: {Linear}-{Time} {Sequence} {Modeling} with {Selective} {State} {Spaces}},
	shorttitle = {Mamba},
	url = {https://openreview.net/forum?id=tEYskw1VY2},
	language = {en},
	urldate = {2026-09-02},
	author = {Gu, Albert and Dao, Tri},
	year = {2024},
}

@inproceedings{graphmixer,
	title = {Do {We} {Really} {Need} {Complicated} {Model} {Architectures} {For} {Temporal} {Networks}?},
	url = {https://openreview.net/forum?id=ayPPc0SyLv1},
	language = {en},
	urldate = {2026-09-02},
	author = {Cong, Weilin and Zhang, Si and Kang, Jian and Yuan, Baichuan and Wu, Hao and Zhou, Xin and Tong, Hanghang and Mahdavi, Mehrdad},
	year = {2023},
}

@inproceedings{rendle2012bprbayesianpersonalizedranking,
	address = {Arlington, Virginia, USA},
	series = {{UAI} '09},
	title = {{BPR}: {Bayesian} personalized ranking from implicit feedback},
	isbn = {978-0-9749039-5-8},
	shorttitle = {{BPR}},
	url = {https://dl.acm.org/doi/10.5555/1795114.1795167},
	urldate = {2026-09-11},
	booktitle = {Proceedings of the {Twenty}-{Fifth} {Conference} on {Uncertainty} in {Artificial} {Intelligence}},
	publisher = {AUAI Press},
	author = {Rendle, Steffen and Freudenthaler, Christoph and Gantner, Zeno and Schmidt-Thieme, Lars},
	month = jun,
	year = {2009},
	pages = {452--461},
}

@article{sorensen1987real,
	title = {Real-valued fast {Fourier} transform algorithms},
	volume = {35},
	copyright = {https://ieeexplore.ieee.org/Xplorehelp/downloads/license-information/IEEE.html},
	issn = {0096-3518},
	url = {http://ieeexplore.ieee.org/document/1165220/},
	doi = {10.1109/TASSP.1987.1165220},
	language = {en},
	number = {6},
	urldate = {2026-09-11},
	journal = {IEEE Transactions on Acoustics, Speech, and Signal Processing},
	author = {Sorensen, H. and Jones, D. and Heideman, M. and Burrus, C.},
	month = jun,
	year = {1987},
	pages = {849--863},
}

@misc{liu2024mamba4rec,
	title = {{Mamba4Rec}: {Towards} {Efficient} {Sequential} {Recommendation} with {Selective} {State} {Space} {Models}},
	shorttitle = {{Mamba4Rec}},
	url = {http://arxiv.org/abs/2403.03900},
	doi = {10.48550/arXiv.2403.03900},
	urldate = {2026-09-09},
	publisher = {arXiv},
	author = {Liu, Chengkai and Lin, Jianghao and Wang, Jianling and Liu, Hanzhou and Caverlee, James},
	month = jun,
	year = {2024},
	note = {arXiv:2403.03900 [cs.IR]},
}

@article{kang2018selfattentive,
	address = {Singapore},
	title = {Self-{Attentive} {Sequential} {Recommendation}},
	url = {https://ieeexplore.ieee.org/document/8594844/},
	doi = {10.1109/ICDM.2018.00035},
	urldate = {2026-09-09},
	journal = {2018 IEEE International Conference on Data Mining (ICDM)},
	publisher = {IEEE},
	author = {Kang, Wang-Cheng and McAuley, Julian},
	month = nov,
	year = {2018},
	note = {Conference Name: 2018 IEEE International Conference on Data Mining (ICDM)
ISBN: 9781538691595},
	pages = {197--206},
}

@misc{yu2023empiricalevaluationtemporalgraph,
	title = {An {Empirical} {Evaluation} of {Temporal} {Graph} {Benchmark}},
	url = {https://arxiv.org/abs/2307.12510v5},
	language = {en},
	urldate = {2026-09-07},
	journal = {arXiv.org},
	author = {Yu, Le},
	month = jul,
	year = {2023},
}

@inproceedings{kim2025lost,
	address = {New York, NY, USA},
	series = {{KDD} '25},
	title = {Lost in {Sequence}: {Do} {Large} {Language} {Models} {Understand} {Sequential} {Recommendation}?},
	isbn = {979-8-4007-1454-2},
	shorttitle = {Lost in {Sequence}},
	url = {https://dl.acm.org/doi/10.1145/3711896.3737035},
	doi = {10.1145/3711896.3737035},
	urldate = {2026-09-04},
	booktitle = {Proceedings of the 31st {ACM} {SIGKDD} {Conference} on {Knowledge} {Discovery} and {Data} {Mining} {V}.2},
	publisher = {Association for Computing Machinery},
	author = {Kim, Sein and Kang, Hongseok and Kim, Kibum and Kim, Jiwan and Kim, Donghyun and Yang, Minchul and Oh, Kwangjin and McAuley, Julian and Park, Chanyoung},
	month = aug,
	year = {2025},
	pages = {1160--1171},
}

@article{Alvarez_Rodriguez_2021,
	title = {Evolutionary dynamics of higher-order interactions in social networks},
	volume = {5},
	issn = {2397-3374},
	url = {https://www.nature.com/articles/s41562-020-01024-1},
	doi = {10.1038/s41562-020-01024-1},
	language = {en},
	number = {5},
	urldate = {2026-09-02},
	journal = {Nature Human Behaviour},
	author = {Alvarez-Rodriguez, Unai and Battiston, Federico and De Arruda, Guilherme Ferraz and Moreno, Yamir and Perc, Matjaž and Latora, Vito},
	month = jan,
	year = {2021},
	pages = {586--595},
}

@article{ding2025dygmambaefficientlymodelinglongterm,
	title = {{DyGMamba}: {Efficiently} {Modeling} {Long}-{Term} {Temporal} {Dependency} on {Continuous}-{Time} {Dynamic} {Graphs} with {State} {Space} {Models}},
	issn = {2835-8856},
	shorttitle = {{DyGMamba}},
	url = {https://openreview.net/forum?id=sq5AJvVuha},
	language = {en},
	urldate = {2026-09-02},
	journal = {Transactions on Machine Learning Research},
	author = {Ding, Zifeng and Li, Yifeng and He, Yuan and Norelli, Antonio and Wu, Jingcheng and Tresp, Volker and Bronstein, Michael M. and Ma, Yunpu},
	month = feb,
	year = {2025},
}

@inproceedings{jodie,
	address = {Anchorage AK USA},
	title = {Predicting {Dynamic} {Embedding} {Trajectory} in {Temporal} {Interaction} {Networks}},
	isbn = {978-1-4503-6201-6},
	url = {https://dl.acm.org/doi/10.1145/3292500.3330895},
	doi = {10.1145/3292500.3330895},
	language = {en},
	urldate = {2026-09-02},
	booktitle = {Proceedings of the 25th {ACM} {SIGKDD} {International} {Conference} on {Knowledge} {Discovery} \& {Data} {Mining}},
	publisher = {ACM},
	author = {Kumar, Srijan and Zhang, Xikun and Leskovec, Jure},
	month = jul,
	year = {2019},
	pages = {1269--1278},
}

@inproceedings{li2025dyg-mambacontinuousstatespace,
	address = {San Diego, California, USA and Mexico City, Mexico},
	title = {{DyG}-{Mamba}: {Continuous} {State} {Space} {Modeling} on {Dynamic} {Graphs}},
	isbn = {979-8-3313-3827-5},
	shorttitle = {{DyG}-{Mamba}},
	url = {http://www.proceedings.com/085713-4299.html},
	doi = {10.52202/085713-4299},
	urldate = {2026-09-02},
	booktitle = {Advances in {Neural} {Information} {Processing} {Systems} 38},
	publisher = {Neural Information Processing Systems Foundation, Inc. (NeurIPS)},
	author = {Li, Dongyuan and Tan, Shiyin and Zhang, Ying and Jin, Ming and Pan, Shirui and Okumura, Manabu and Jiang, Renhe},
	year = {2025},
	pages = {143195--143224},
}

@inproceedings{pan2020star,
	address = {Virtual Event Ireland},
	title = {Star {Graph} {Neural} {Networks} for {Session}-based {Recommendation}},
	isbn = {978-1-4503-6859-9},
	url = {https://dl.acm.org/doi/10.1145/3340531.3412014},
	doi = {10.1145/3340531.3412014},
	language = {en},
	urldate = {2026-09-02},
	booktitle = {Proceedings of the 29th {ACM} {International} {Conference} on {Information} \& {Knowledge} {Management}},
	publisher = {ACM},
	author = {Pan, Zhiqiang and Cai, Fei and Chen, Wanyu and Chen, Honghui and De Rijke, Maarten},
	month = oct,
	year = {2020},
	pages = {1195--1204},
}

@misc{tgns,
	title = {Temporal {Graph} {Networks} for {Deep} {Learning} on {Dynamic} {Graphs}},
	url = {http://arxiv.org/abs/2006.10637},
	doi = {10.48550/arXiv.2006.10637},
	urldate = {2026-09-02},
	publisher = {arXiv},
	author = {Rossi, Emanuele and Chamberlain, Ben and Frasca, Fabrizio and Eynard, Davide and Monti, Federico and Bronstein, Michael},
	month = oct,
	year = {2020},
	note = {arXiv:2006.10637 [cs.LG]},
}

@article{tcl,
	title = {{TCL}: {Transformer}-based {Dynamic} {Graph} {Modelling} via {Contrastive} {Learning}},
	shorttitle = {{TCL}},
	url = {https://www.semanticscholar.org/paper/TCL%3A-Transformer-based-Dynamic-Graph-Modelling-via-Wang-Chang/d37a8e9cea2f1e1b45b4912d243eee1b9dd5f756},
	urldate = {2026-09-02},
	journal = {ArXiv},
	author = {Wang, Lu and Chang, Xiaofu and Li, Shuang and Chu, Yunfei and Li, Hui and Zhang, Wei and He, Xiaofeng and Song, Le and Zhou, Jingren and Yang, Hongxia},
	month = may,
	year = {2021},
}

@inproceedings{craft,
	address = {San Diego, California, USA and Mexico City, Mexico},
	title = {Future {Link} {Prediction} {Without} {Memory} or {Aggregation}},
	isbn = {979-8-3313-3827-5},
	url = {http://www.proceedings.com/085713-0084.html},
	doi = {10.52202/085713-0084},
	urldate = {2026-09-02},
	booktitle = {Advances in {Neural} {Information} {Processing} {Systems} 38},
	publisher = {Neural Information Processing Systems Foundation, Inc. (NeurIPS)},
	author = {Yi, Lu and Lei, Runlin and Mo, Fengran and Zheng, Yanping and Wei, Zhewei and Ye, Yuhang},
	year = {2025},
	pages = {2672--2700},
}

@article{yi2025tgbseqbenchmarkchallengingtemporal,
	title = {{TGB}-{Seq} {Benchmark}: {Challenging} {Temporal} {GNNs} with {Complex} {Sequential} {Dynamics}},
	volume = {2025},
	shorttitle = {{TGB}-{Seq} {Benchmark}},
	url = {https://proceedings.iclr.cc/paper_files/paper/2025/hash/db5ca61dbc08cf5143c05ad2d1b0b2ca-Abstract-Conference.html},
	language = {en},
	urldate = {2026-09-02},
	journal = {International Conference on Learning Representations},
	author = {Yi, Lu and Peng, Jie and Zheng, Yanping and Mo, Fengran and Wei, Zhewei and Ye, Yuhang and Zixuan, Yue and Huang, Zengfeng},
	month = may,
	year = {2025},
	pages = {88022--88042},
}

@incollection{craswell2009mrr,
	address = {Boston, MA},
	title = {Mean {Reciprocal} {Rank}},
	isbn = {978-0-387-35544-3 978-0-387-39940-9},
	url = {http://link.springer.com/10.1007/978-0-387-39940-9_488},
	doi = {10.1007/978-0-387-39940-9_488},
	language = {en},
	urldate = {2026-09-02},
	booktitle = {Encyclopedia of {Database} {Systems}},
	publisher = {Springer US},
	author = {Craswell, Nick},
	editor = {Liu, Ling and Özsu, M. Tamer},
	year = {2009},
	pages = {1703--1703},
}

@article{10.1007/s00354-019-00065-z,
	title = {Temporal {Link} {Prediction}: {A} {Survey}},
	volume = {38},
	issn = {0288-3635, 1882-7055},
	shorttitle = {Temporal {Link} {Prediction}},
	url = {http://link.springer.com/10.1007/s00354-019-00065-z},
	doi = {10.1007/s00354-019-00065-z},
	language = {en},
	number = {1},
	urldate = {2026-09-02},
	journal = {New Generation Computing},
	author = {Divakaran, Aswathy and Mohan, Anuraj},
	month = mar,
	year = {2020},
	pages = {213--258},
}

@inproceedings{huang2023temporalgraphbenchmarkmachine,
	address = {New Orleans, Louisiana, USA},
	title = {Temporal {Graph} {Benchmark} for {Machine} {Learning} on {Temporal} {Graphs}},
	isbn = {978-1-7138-9911-2},
	url = {http://www.proceedings.com/075280-0099.html},
	doi = {10.52202/075280-0099},
	urldate = {2026-09-02},
	booktitle = {Advances in {Neural} {Information} {Processing} {Systems} 36},
	publisher = {Neural Information Processing Systems Foundation, Inc. (NeurIPS)},
	author = {Huang, Shenyang and Poursafaei, Farimah and Danovitch, Jacob and Fey, Matthias and Hu, Weihua and Rossi, Emanuele and Leskovec, Jure and Bronstein, Michael and Rabusseau, Guillaume and Rabbany, Reihaneh},
	year = {2023},
	pages = {2056--2073},
}

@inproceedings{SimpleDyG,
	address = {Singapore Singapore},
	title = {On the {Feasibility} of {Simple} {Transformer} for {Dynamic} {Graph} {Modeling}},
	isbn = {979-8-4007-0171-9},
	url = {https://dl.acm.org/doi/10.1145/3589334.3645622},
	doi = {10.1145/3589334.3645622},
	language = {en},
	urldate = {2026-09-02},
	booktitle = {Proceedings of the {ACM} {Web} {Conference} 2024},
	publisher = {ACM},
	author = {Wu, Yuxia and Fang, Yuan and Liao, Lizi},
	month = may,
	year = {2024},
	pages = {870--880},
}

@article{yu2023continuoustimeuserpreferencemodelling,
	title = {Continuous-{Time} {User} {Preference} {Modelling} for {Temporal} {Sets} {Prediction}},
	volume = {36},
	copyright = {https://ieeexplore.ieee.org/Xplorehelp/downloads/license-information/IEEE.html},
	issn = {1041-4347, 1558-2191, 2326-3865},
	url = {https://ieeexplore.ieee.org/document/10234655/},
	doi = {10.1109/TKDE.2023.3309982},
	number = {4},
	urldate = {2026-09-02},
	journal = {IEEE Transactions on Knowledge and Data Engineering},
	author = {Yu, Le and Liu, Zihang and Sun, Leilei and Du, Bowen and Liu, Chuanren and Lv, Weifeng},
	month = apr,
	year = {2024},
	pages = {1475--1488},
}

@inproceedings{dygformer,
	address = {New Orleans, Louisiana, USA},
	title = {Towards {Better} {Dynamic} {Graph} {Learning}: {New} {Architecture} and {Unified} {Library}},
	isbn = {978-1-7138-9911-2},
	shorttitle = {Towards {Better} {Dynamic} {Graph} {Learning}},
	url = {http://www.proceedings.com/075280-2960.html},
	doi = {10.52202/075280-2960},
	urldate = {2026-09-02},
	booktitle = {Advances in {Neural} {Information} {Processing} {Systems} 36},
	publisher = {Neural Information Processing Systems Foundation, Inc. (NeurIPS)},
	author = {Yu, Le and Sun, Leilei and Du, Bowen and Lv, Weifeng},
	year = {2023},
	pages = {67686--67700},
}
\bibliographystyle{iclr2027_conference}

\clearpage
\appendix
\appendix

\section{Abbreviations}
\label{app:abbreviations}

\begin{table}[htbp]
\centering
\caption{Key abbreviations and shorthand.}
\label{tab:abbreviations}
\small
\begin{tabular}{@{}ll@{}}
\hline
\textbf{Term} & \textbf{Meaning} \\
\hline
CAME & Candidate-Aware Mamba Encoder \\
DRR & Dual-Branch Representation Refinement \\
HSR & Historical Spectral Refinement \\
CFR & Candidate Feature Refinement \\
MRR & Mean Reciprocal Rank \\
-R & Suffix indicating historical repetition frequency \\
w/o & Without the specified component \\
pp & Percentage points \\
OOM & Exceeds the memory budget \\
OOT & Exceeds the time budget \\
\hline
\end{tabular}
\end{table}

\section{Experimental Details}
\subsection{Hardware and Software Environment}
All experiments were conducted on NVIDIA RTX PRO 6000 Blackwell GPUs with 96GB of memory. Our implementation uses Python~3.11, PyTorch~2.7.1, and CUDA~12.8, together with \texttt{mamba-ssm}~2.3.1 and \texttt{causal-conv1d}~1.6.2. 

\subsection{Dataset Details}
\label{app:datasets}
Our main evaluation includes 13 future link prediction datasets. Seven are from TGB-Seq~\citep{yi2025tgbseqbenchmarkchallengingtemporal}: ML-20M, Taobao, Yelp, GoogleLocal, Flickr, YouTube, and WikiLink. The remaining six are commonly used temporal graph datasets: Wikipedia, Reddit, MOOC, LastFM, UCI, and Flights. We additionally evaluate MambaTG on three datasets from TGB~\citep{huang2023temporalgraphbenchmarkmachine}: \texttt{tgbl-comment}, \texttt{tgbl-coin}, and \texttt{tgbl-flight}. These additional results are reported in Appendix~\ref{app:additional_results_tgbl_datasets}.

Following \citet{craft}, we classify datasets as \emph{unseen-edge-dominant} if most target interactions are previously unseen edges, and as \emph{repeated-edge-dominant} if most are repeated edges. The datasets are obtained from the \href{https://huggingface.co/TGB-Seq}{TGB-Seq collection}, the \href{https://tgb.complexdatalab.com/}{TGB repository}, and the \href{https://zenodo.org/records/7213796}{public Zenodo archive}. Table~\ref{tab:dataset_statistics} summarizes their statistics; for bipartite datasets, the two node counts correspond to the source and destination node types.

\subsection{Baseline Details}
\label{app:baseline_details}
We compare MambaTG with established temporal graph models, recent Mamba-based architectures, a sequential recommendation method, and a candidate-aware future link predictor. Their main designs are summarized below:

\paragraph{EdgeBank.}
EdgeBank~\citep{poursafaei2022towards} is a non-parametric baseline that stores previously observed edges and scores candidate links according to whether they have appeared in the interaction history. It captures edge recurrence without learning node representations or model parameters. Given its explicit focus on edge recurrence, we report EdgeBank only on repeated-edge-dominant datasets.

\paragraph{JODIE.}
JODIE~\citep{jodie} maintains dynamic source and destination embeddings using two coupled recurrent neural networks. After each interaction, it jointly updates the embeddings of both endpoints and projects the source embedding forward to the query time for future interaction prediction.
\paragraph{DyRep.}
DyRep~\citep{DyRep} models communication and association events as two interacting temporal processes. It updates persistent node representations using temporally attentive neighborhood information and parameterizes event occurrence through a temporal point process.
\paragraph{TGAT.}
TGAT~\citep{tgat} computes time-dependent node representations by applying self-attention over sampled temporal neighborhoods. It uses a functional time encoding to represent the elapsed time of historical interactions and supports inductive inference without maintaining persistent node memories.
\paragraph{TGN.}
TGN~\citep{tgns} maintains a persistent memory for each node. Incoming interactions generate messages that are aggregated and used to update the endpoint memories, after which a graph-based embedding module combines the memories with temporal-neighborhood information for prediction.
\paragraph{CAWN.}
CAWN~\citep{cawn} samples causal temporal walks from the two endpoints of a query edge. It anonymizes node identities through walk-based occurrence patterns and encodes the resulting walks to capture temporal motifs and structural relationships relevant to the queried pair.
\paragraph{GraphMixer.}
GraphMixer~\citep{graphmixer} uses an MLP-Mixer-based link encoder to summarize recent temporal interactions, a mean-pooling node encoder to aggregate node features, and an MLP classifier for link prediction. It provides a memory-free alternative to recurrent and attention-based temporal graph models.
\paragraph{DyGFormer.}
DyGFormer~\citep{dygformer} extracts the first-hop interaction histories of both endpoints and uses neighbor co-occurrence encoding to represent correlations between them. The histories are divided into patches and processed by a Transformer to capture long-range temporal dependencies efficiently.

\paragraph{FreeDyG.}
FreeDyG~\citep{tian2024freedyg} is a frequency-enhanced continuous-time dynamic graph model for link prediction. It introduces node interaction frequency encoding to explicitly capture the proportion of common neighbors and the interaction frequency of each node pair. By modeling interaction patterns in the frequency domain, FreeDyG aims to capture periodic behavior and temporal shifts in node interactions.

\paragraph{DyGMamba.}
DyGMamba~\citep{ding2025dygmambaefficientlymodelinglongterm} employs two state-space components. A node-level SSM encodes historical node interactions, while a time-level SSM captures temporal patterns and dynamically selects relevant information from the extended interaction history.
\paragraph{DyG-Mamba.}
DyG-Mamba~\citep{li2025dyg-mambacontinuousstatespace} formulates dynamic graph learning using a continuous state-space architecture inspired by Mamba. It uses irregular time intervals as control signals for the state transitions, allowing the model to adjust how historical information is retained or forgotten.
\paragraph{SGNN-HN.}
SGNN-HN~\citep{pan2020star} was originally proposed for session-based recommendation. It represents an ordered interaction sequence as a star graph, allowing the star node to exchange information with all items in the sequence. A highway network combines item representations before and after graph propagation, and an attention mechanism aggregates them into a sequence representation for candidate prediction.

\paragraph{SASRec.}
SASRec~\citep{kang2018selfattentive} is a self-attention-based sequential recommendation model that encodes a user's ordered item-interaction history to predict the next item, capturing dependencies among past interactions.

\paragraph{Mamba4Rec.}
Mamba4Rec~\citep{liu2024mamba4rec} is a Mamba-based sequential recommendation model that encodes a user's ordered item-interaction history for next-item prediction through selective state-space updates.

\paragraph{CRAFT.}
CRAFT~\citep{craft} represents nodes using trainable identifier embeddings and applies cross-attention from each candidate destination to the source node's recent interactions. It therefore performs direct candidate-to-history matching without maintaining persistent node memories or aggregating the candidate's temporal neighborhood. Following its original design, the repetition-enhanced CRAFT-R variant additionally supplies historical repetition frequency to the predictor on repeated-edge-dominant datasets.

\begin{table*}[h]
\centering
\caption{Statistics of the 16 datasets used in our experiments. Regime indicates whether a dataset is unseen-edge-dominant (Unseen) or repeated-edge-dominant (Repeated).}
\label{tab:dataset_statistics}
\resizebox{\textwidth}{!}{
\begin{tabular}{lrrrrrcll}
\toprule
\textbf{Dataset} & \textbf{Nodes} & \textbf{Edges} & \textbf{Timestamps} & \textbf{Repeat Ratio (\%)} & \textbf{Density (\%)} & \textbf{Bipartite} & \textbf{Domain} & \textbf{Regime} \\
\midrule
ML-20M       & 100,785/9,646     & 14,494,325 & 9,993,250  & 0     & $1.49 \times 10^{0}$  & Yes & Movie rating           & Unseen \\
Taobao       & 760,617/863,016   & 18,853,792 & 139,171    & 16.58 & $2.87 \times 10^{-3}$ & Yes & E-commerce interaction & Unseen \\
Yelp         & 1,338,688/405,081 & 19,760,293 & 14,646,734 & 25.18 & $3.64 \times 10^{-3}$ & Yes & Business review        & Unseen \\
GoogleLocal  & 206,244/267,336   & 1,913,967  & 1,771,060  & 0     & $3.47 \times 10^{-3}$ & Yes & Business review        & Unseen \\
Flickr       & 233,836           & 7,223,559  & 134        & 0     & $1.32 \times 10^{-2}$ & No  & Who-to-follow          & Unseen \\
YouTube      & 402,422           & 3,288,028  & 203        & 0     & $2.03 \times 10^{-3}$ & No  & Who-to-follow          & Unseen \\
WikiLink     & 1,361,972         & 34,163,774 & 2,198      & 0     & $1.84 \times 10^{-3}$ & No  & Web link               & Unseen \\
\midrule
Wikipedia    & 8,227/1,000       & 157,474    & 152,757    & 88.41 & $1.91 \times 10^{0}$  & Yes & Interaction            & Repeated \\
Reddit       & 10,000/984        & 672,447    & 669,065    & 88.32 & $6.83 \times 10^{0}$  & Yes & Social                 & Repeated \\
MOOC         & 7,047/97          & 411,749    & 345,600    & 56.66 & $6.02 \times 10^{1}$  & Yes & Interaction            & Repeated \\
LastFM       & 980/1,000         & 1,293,103  & 1,283,614  & 88.01 & $1.32 \times 10^{2}$  & Yes & Interaction            & Repeated \\
UCI          & 1,899             & 59,835     & 58,911     & 66.06 & $1.66 \times 10^{0}$  & No  & Social                 & Repeated \\
Flights      & 13,169            & 1,927,145  & 122        & 79.50 & $1.11 \times 10^{0}$  & No  & Transport              & Repeated \\
\midrule
tgbl-comment & 994,790           & 44,314,507 & 30,998,030 & 19.81 & $4.48 \times 10^{-3}$ & No  & Social                 & Unseen \\
tgbl-coin    & 638,486           & 22,809,486 & 1,295,720  & 82.93 & $5.60 \times 10^{-3}$ & No  & Transaction            & Repeated \\
tgbl-flight  & 18,143            & 67,169,570 & 1,385      & 96.48 & $2.04 \times 10^{1}$  & No  & Transport              & Repeated \\
\bottomrule
\end{tabular}
}
\end{table*}

\subsection{Evaluation and Implementation Details}
\label{app:implementation_details}

Following TGB-Seq~\citep{yi2025tgbseqbenchmarkchallengingtemporal} and DyGLib~\citep{dygformer}, we use the Adam optimizer with a learning rate of \(10^{-4}\) and a batch size of 200, which is increased to 400 for large datasets. Models are trained for up to 100 epochs with early stopping using a patience of 20 epochs. We impose a computational budget of 1.5 hours per epoch and 32\,GB of GPU memory; exceeding these limits is denoted as out of time (OOT) and out of memory (OOM), respectively. We use official negative samples whenever available. TGB~\citep{huang2023temporalgraphbenchmarkmachine} provides either 20 or 100 negative destinations per positive edge, depending on the dataset, while TGB-Seq provides 100 randomly sampled negative destinations. For datasets without official negative samples, we follow TGB-Seq and randomly sample 100 negative destinations per positive edge. For small datasets, we additionally follow CRAFT~\citep{craft} and perform collision checks to exclude positive edges from the sampled negatives. To ensure fair comparison, we use published baseline results when they follow the same evaluation protocol and closely matched training settings; otherwise, we reproduce the methods using our unified pipeline.

\subsection{Hyperparameter Settings}
\label{app:hyperparameter_settings}

\paragraph{Baseline configurations.}
For each baseline we reproduce, we use the best configuration reported for the corresponding baseline--dataset pair in the original paper or benchmark study, including TGB-Seq~\citep{yi2025tgbseqbenchmarkchallengingtemporal}, CRAFT~\citep{craft}, or DyGLib~\citep{dygformer}. When no dataset-specific configuration is reported, we use the default configuration recommended by the baseline's official implementation.

\paragraph{MambaTG configurations.}
We select MambaTG hyperparameters according to validation MRR. Specifically, we tune the maximum history length \(L_{\max}\), number of CAME layers \(R\), hidden-layer dropout rate \(p_{\mathrm{h}}\), and node-embedding dropout rate \(p_{\mathrm{e}}\) using the search space in Table~\ref{tab:tuning_hyperparameters}. Across all datasets, we fix the Mamba state dimension to \(d_{\mathrm{state}}=32\), the expansion factor to 2, and the causal-convolution width to 4. Table~\ref{tab:final_hyperparameters} reports the selected configurations, where \(d_e\) denotes the node-embedding dimension and \(B\) the training batch size. Following \citet{craft}, we adopt the same dataset-specific settings for the node-embedding dimension \(d_e\) and training batch size \(B\) to ensure a fair comparison.

\begin{table}[h]
\centering
\caption{Hyperparameter search space for MambaTG.}
\label{tab:tuning_hyperparameters}
\begin{tabular}{ll}
\toprule
\textbf{Hyperparameter} & \textbf{Search space} \\
\midrule
Maximum history length, \(L_{\max}\)
    & \(\{30, 60, 90, 120, 250\}\) \\
Number of CAME layers, \(R\)
    & \(\{1, 2\}\) \\
Hidden-layer dropout, \(p_{\mathrm{h}}\)
    & \(\{0.1, 0.2\}\) \\
Node-embedding dropout, \(p_{\mathrm{e}}\)
    & \(\{0.1, 0.2\}\) \\
\bottomrule
\end{tabular}
\end{table}

\begin{table*}[h]
\centering
\caption{Selected MambaTG hyperparameters for each dataset.}
\label{tab:final_hyperparameters}
\begin{tabular}{lcccccc}
\toprule
\textbf{Dataset} &
\(\boldsymbol{L_{\max}}\) &
\(\boldsymbol{R}\) &
\(\boldsymbol{p_{\mathrm{h}}}\) &
\(\boldsymbol{p_{\mathrm{e}}}\) &
\(\boldsymbol{d_e}\) &
\(\boldsymbol{B}\) \\
\midrule
UCI          & 30  & 1 & 0.1 & 0.1 & 64  & 200 \\
Wikipedia    & 90 & 2 & 0.2 & 0.2 & 64  & 200 \\
Reddit       & 120 & 1 & 0.1 & 0.1 & 64  & 200 \\
MOOC         & 30  & 2 & 0.1 & 0.1 & 64  & 200 \\
Flights      & 250 & 1 & 0.1 & 0.1 & 64  & 200 \\
LastFM       & 250 & 2 & 0.1 & 0.1 & 64  & 200 \\
\midrule
GoogleLocal  & 250 & 2 & 0.1 & 0.2 & 128 & 200 \\
Flickr       & 90  & 1 & 0.1 & 0.1 & 128 & 400 \\
YouTube      & 120 & 2 & 0.1 & 0.1 & 128 & 400 \\
Taobao       & 60  & 2 & 0.1 & 0.1 & 128 & 400 \\
Yelp         & 60  & 2 & 0.1 & 0.1 & 128 & 400 \\
ML-20M       & 60  & 1 & 0.1 & 0.1 & 128 & 400 \\
WikiLink     & 250 & 1 & 0.1 & 0.1 & 128 & 400 \\
\midrule
tgbl-comment & 30  & 1 & 0.1 & 0.1 & 128 & 200 \\
tgbl-coin    & 250  & 1 & 0.1 & 0.1 & 128 & 200 \\
tgbl-flight  & 90  & 1 & 0.1 & 0.1 & 128 & 200 \\
\bottomrule
\end{tabular}
\end{table*}

\section{Method Details}
\subsection{Mamba Block and Feed-Forward Network}
\label{app:mamba_block}

\paragraph{Mamba block.}
Each Mamba block takes an input sequence \(\mathbf{X}\in\mathbb{R}^{m\times d}\). Following \citet{gu2024mambalineartimesequencemodeling}, it applies an input projection and a causal one-dimensional convolution, processes the resulting sequence with a selective SSM, and gates the SSM output:
\begin{equation}
\begin{gathered}
\mathbf{U}
=
\operatorname{SiLU}\!\left(
\operatorname{Conv1D}\!\left(
\operatorname{Linear}_{\mathrm{in}}(\mathbf{X})
\right)
\right),
\qquad
\mathbf{Y}
=
\operatorname{SSM}_{\overline{\mathbf{A}},
\overline{\mathbf{B}},\mathbf{C}}(\mathbf{U}),\\
\operatorname{MambaBlock}(\mathbf{X})
=
\operatorname{Linear}_{\mathrm{out}}\!\left(
\mathbf{Y}\odot
\operatorname{SiLU}\!\left(
\operatorname{Linear}_{\mathrm{gate}}(\mathbf{X})
\right)
\right).
\end{gathered}
\end{equation}

The selective SSM processes \(\mathbf{U}\) recurrently. At position \(t\), it maps \(\mathbf{u}_t\) to \(\mathbf{y}_t\) through the hidden state \(\mathbf{h}_t\):
\begin{equation}
\mathbf{h}_t
=
\overline{\mathbf{A}}_t\mathbf{h}_{t-1}
+
\overline{\mathbf{B}}_t\mathbf{u}_t,
\qquad
\mathbf{y}_t
=
\mathbf{C}_t\mathbf{h}_t.
\end{equation}
Here, \(\overline{\mathbf{A}}_t\) controls how the preceding state is retained, \(\overline{\mathbf{B}}_t\) determines how the current input updates the state, and \(\mathbf{C}_t\) maps the hidden state to the output. Mamba derives \(\mathbf{B}_t\), \(\mathbf{C}_t\), and the discretization step \(\boldsymbol{\Delta}_t\) from the current input:
\begin{equation}
\mathbf{B}_t
=
\mathbf{u}_t\mathbf{W}_{B},
\qquad
\mathbf{C}_t
=
\mathbf{u}_t\mathbf{W}_{C},
\qquad
\boldsymbol{\Delta}_t
=
\operatorname{softplus}\!\left(
\operatorname{Broadcast}_{d_{\mathrm{inner}}}
\left(\mathbf{u}_t\mathbf{W}_{\Delta}\right)
+
\mathbf{b}_{\Delta}
\right).
\end{equation}
Under zero-order-hold discretization, the shared continuous state-transition parameter \(\mathbf{A}\) and input-dependent parameter \(\mathbf{B}_t\) are converted into their discrete forms:
\begin{equation}
\overline{\mathbf{A}}_t
=
\exp(\boldsymbol{\Delta}_t\mathbf{A}),
\qquad
\overline{\mathbf{B}}_t
=
(\boldsymbol{\Delta}_t\mathbf{A})^{-1}
\left[
\exp(\boldsymbol{\Delta}_t\mathbf{A})-\mathbf{I}
\right]
\boldsymbol{\Delta}_t\mathbf{B}_t.
\end{equation}

\paragraph{Feed-forward network.}
Each Mamba block is followed by a position-wise FFN:
\begin{equation}
\operatorname{FFN}(\mathbf{X})
=
\operatorname{GELU}\!\left(
\mathbf{X}\mathbf{W}_1+\mathbf{b}_1
\right)\mathbf{W}_2+\mathbf{b}_2,
\end{equation}
where \(\mathbf{W}_1\in\mathbb{R}^{d\times4d}\) and \(\mathbf{W}_2\in\mathbb{R}^{4d\times d}\).

\subsection{CAME implementation.}
\label{app:came_implementation}

At a given query time, CAME encodes each source history once and retains its layer-wise terminal recurrent caches \(K_H^{(r)}\) for candidate continuation during both training and inference. Each candidate then continues independently from the same terminal history caches, so candidates share the encoded history without affecting one another.

For training, however, the terminal SSM state returned by the standard \texttt{mamba-ssm} selective-scan interface does not support gradient propagation. To allow the BPR objective to propagate gradients through candidate continuation to the history encoder, we extend each history scan from length \(L_{\max}\) to \(L_{\max}+d_{\mathrm{state}}\), and append \(d_{\mathrm{state}}\) auxiliary positions to expose the terminal SSM state through the scan outputs. For each auxiliary position \(t=L_{\max}+j\), we set \(\mathbf{u}_t=\mathbf{0}\), \(\mathbf{B}_t=\mathbf{0}\), and \(\widetilde{\boldsymbol{\Delta}}_t=\Delta_{\mathrm{off}}\mathbf{1}\), where \(\Delta_{\mathrm{off}}\) is chosen such that \(\widetilde{\boldsymbol{\Delta}}_t+\mathbf{b}_{\Delta}\) is componentwise sufficiently negative. Thus, \(\boldsymbol{\Delta}_t=\operatorname{softplus}(\widetilde{\boldsymbol{\Delta}}_t+\mathbf{b}_{\Delta})\approx\mathbf{0}\), giving \(\overline{\mathbf{A}}_t=\exp(\boldsymbol{\Delta}_t\mathbf{A})\approx\mathbf{I}\) and \(\overline{\mathbf{B}}_t\mathbf{u}_t=\mathbf{0}\), and hence \(\mathbf{h}_{L_{\max}+j}\approx\mathbf{h}_{L_{\max}}\). Although the softplus parameterization keeps \(\boldsymbol{\Delta}_t>0\), the sufficiently negative \(\Delta_{\mathrm{off}}\) makes the resulting state decay numerically negligible. We then read out the terminal state from the auxiliary outputs by setting \(\mathbf{C}_{L_{\max}+j}=\mathbf{e}_j^\top\) at auxiliary position \(t=L_{\max}+j\), where \(\mathbf{e}_j\in\mathbb{R}^{d_{\mathrm{state}}}\) is the \(j\)-th standard basis vector. Since \(\mathbf{u}_{L_{\max}+j}=\mathbf{0}\) and \(\mathbf{h}_{L_{\max}+j}\approx\mathbf{h}_{L_{\max}}\):
\begin{equation}
\mathbf{y}_{L_{\max}+j}
=
\mathbf{C}_{L_{\max}+j}\mathbf{h}_{L_{\max}+j}
\approx
\mathbf{e}_j^\top\mathbf{h}_{L_{\max}}.
\end{equation}
The \(d_{\mathrm{state}}\) auxiliary outputs therefore jointly recover the terminal SSM state through the standard selective-scan interface, without modifying its CUDA kernel. The SiLU gate is applied only to the original history outputs \(\mathbf{y}_{1:L_{\max}}\), while the auxiliary outputs are used directly for state readout. 

For batched histories with different true lengths, each history is right-padded to \(L_{\max}\). At padded positions, we set \(\mathbf{u}_t=\mathbf{0}\), \(\mathbf{B}_t=\mathbf{0}\), and \(\tilde{\boldsymbol{\Delta}}_t=\Delta_{\mathrm{off}}\mathbf{1}\), following the approximate-identity construction above. Consequently, the scan can operate on the common length \(L_{\max}\) without introducing state decay after the last valid interaction. The terminal causal-convolution state is gathered from the last valid positions of each history rather than from the padded positions.

\section{Additional Details and Results for Signal Ablations}
\label{app:signal_ablation_details}

We provide the experimental settings for ShuffledTrain, ShuffledInference, ShuffledBoth, and LateFusion, which examine \emph{historical-sequence evolution} and \emph{candidate--history compatibility}. Unless otherwise stated, all experiments follow the same data splits, historical sampling strategy, negative samples, model configuration, optimization settings, and evaluation protocol as MambaTG.

\subsection{ShuffledTrain}

ShuffledTrain examines how disrupting chronological order during training affects predictive performance. We randomly permute the source node's historical interactions for every training query. Each permutation is generated once before training and remains fixed across epochs, preventing repeated shuffling from acting as stochastic data augmentation. All attributes associated with an interaction, including its timestamp, are permuted together, while the candidate remains the final input token. The model is trained from scratch on these shuffled histories and evaluated using the original chronologically ordered test histories. This variant probes the role of learning from ordered interactions in capturing \emph{historical-sequence evolution}.

\subsection{ShuffledInference}

ShuffledInference examines whether the trained model relies on chronological ordering when making predictions. We train MambaTG normally and randomly permute the source histories only during test-time inference, without retraining or modifying its parameters. As in ShuffledTrain, all attributes associated with each interaction are permuted together, and the candidate remains the final token. A performance reduction relative to standard chronological inference indicates that the model's predictions are sensitive to historical order.

\subsection{ShuffledBoth}
ShuffledBoth examines the benefit of chronological ordering under matched training and evaluation ordering conditions. Since ShuffledTrain and ShuffledInference use different ordering conditions during training and evaluation, their performance changes may also reflect sensitivity to this mismatch. To address this, ShuffledBoth uses shuffled histories during both training and evaluation. 

The model is trained from scratch following the ShuffledTrain procedure and evaluated on shuffled test histories. This comparison tests whether the chronological order of a source node's interaction history provides predictive information for future link prediction beyond historical-neighbor identities.

\subsection{LateFusion}
LateFusion examines whether candidate--history compatibility is better captured by incorporating the candidate into the state transition rather than combining independently encoded representations at the prediction stage. In MambaTG, the complete history encoder processes the historical sequence $H_u(t)$ and retains the layer-wise history caches. The candidate then continues through all $R$ CAME layers using these caches, following Equation~(6):
\begin{equation}
\begin{aligned}
(\mathbf{h}_H,\mathcal{K}_H)
&= \operatorname{HistoryEncoder}\!\left(H_u(t)\right),\\
\mathbf{h}_c
&= \operatorname{CandidateContinuation}\!\left(
\mathbf{z}_{H,c}^{\text{CFR}};\mathcal{K}_H
\right),\\
\hat{y}_{u,c}
&= \operatorname{MLP}\!\left(\mathbf{h}_c\right).
\end{aligned}
\end{equation}
Here, $\mathbf{h}_H$ denotes the final history representation, $\mathcal{K}_H$ contains the recurrent and convolutional states retained at each layer, and $\mathbf{z}_{H,c}^{\text{CFR}}$ denotes the refined candidate representation.

LateFusion removes this candidate continuation while keeping the complete chronological history encoder unchanged. Instead, it independently refines the candidate with a residual MLP and combines it with the history representation only at the prediction stage:
\begin{equation}
\begin{aligned}
\mathbf{h}_H
&= \operatorname{HistoryEncoder}\!\left(H_u(t)\right),\\
\widetilde{\mathbf{z}}_{H,c}^{\text{CFR}}
&= \operatorname{ResMLP}\!\left(
\mathbf{z}_{H,c}^{\text{CFR}}
\right),\\
\hat{y}_{u,c}
&= \operatorname{MLP}\!\left(
[\mathbf{h}_H \,\|\, \widetilde{\mathbf{z}}_{H,c}^{\text{CFR}}]
\right).
\end{aligned}
\end{equation}
For LateFusion, only the history representation is used at the prediction stage, and the history caches are not used for candidate processing. The residual MLP is sized to approximately match the parameter count of one CAME layer. MambaTG reuses its history-encoder parameters for candidate continuation, while LateFusion adds a separate residual MLP and uses a wider prediction head. As a result, LateFusion has more total trainable parameters. Thus, the comparison evaluates candidate-conditioned continuation against late fusion with a larger parameter budget, ensuring that any advantage of MambaTG cannot be attributed simply to a higher total parameter count. LateFusion is trained from scratch using the same training and evaluation settings as MambaTG. The explicit features used by the MambaTG link classifier are retained but omitted from both equations for clarity.

\subsection{Per-Dataset Results}
Figure~\ref{fig:cand_hist_compat_all_datasets} reports the per-dataset results for all four signal ablations.

\begin{figure}[h]
    \centering
    \includegraphics[width=\linewidth]{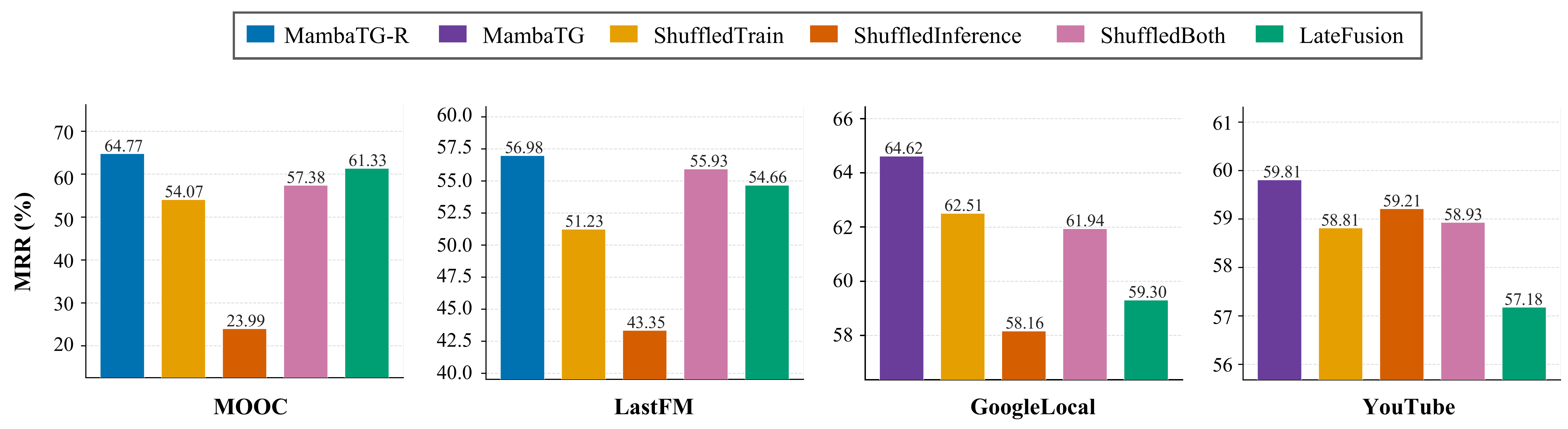}
    \caption{Per-dataset signal-ablation results on the repeated-edge-dominant datasets (MOOC and LastFM) and the unseen-edge-dominant datasets (GoogleLocal and YouTube). Results are reported as MRR (\%), averaged over three random seeds.}
    \label{fig:cand_hist_compat_all_datasets}
\end{figure}

\section{Ablation Study Details}
\label{app:component_ablation}

We conduct an ablation study on four datasets to evaluate the contributions of MambaTG's key components. Specifically, we consider five variants: (1) \textbf{w/o HSR}, which removes the Historical Spectral Refinement (HSR) applied to the source node's historical interaction sequence; (2) \textbf{w/o CFR}, which removes Candidate Feature Refinement (CFR), preventing the candidate representations from being adaptively aligned with the refined historical representation space; (3) \textbf{w/o query-rel}, which removes the query-relative time encoding from the input representations; (4) \textbf{w/o recency}, which removes the candidate activity recency feature concatenated at the prediction layer; and (5) \textbf{w/o repeat}, which removes the candidate repetition-frequency feature from the prediction layer. The repetition-frequency feature is evaluated only on repeated-edge datasets. Figure~\ref{fig:mambatg_ablation} presents the complete per-dataset results.

\begin{figure}[h]
\centering
\includegraphics[width=\linewidth]{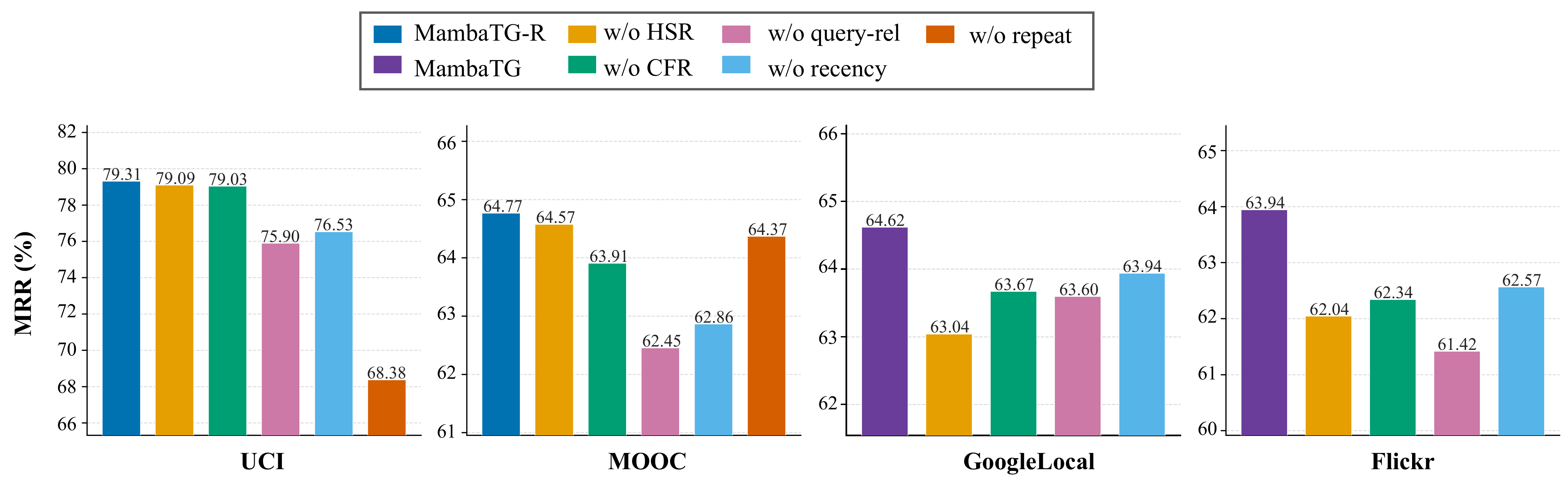}
\caption{Per-dataset component-ablation results on the repeated-edge-dominant datasets (UCI and MOOC) and the unseen-edge-dominant datasets (GoogleLocal and Flickr). MambaTG-R denotes MambaTG augmented with historical repetition frequency. Each ablation variant removes one component from the corresponding complete model. Results are reported as MRR (\%), averaged over three random seeds.}
\label{fig:mambatg_ablation}
\end{figure}

\section{Effectiveness and Transferability of Dual-Branch Representation Refinement (DRR)}
\label{app:drr_impact}

DRR improves the inputs to candidate-aware temporal modeling through two complementary branches: HSR refines historical patterns at different temporal scales, while CFR recalibrates candidate features to promote compatibility with the refined history. To examine whether its benefits extend beyond MambaTG, we incorporate DRR into CRAFT and compare the resulting \textbf{CRAFT-DRR} with the original CRAFT. Specifically, we construct CRAFT-DRR by applying HSR to the history embeddings after positional encoding and CFR to the candidate embeddings, with the resulting representations passed to CRAFT's cross-attention module. We also compare \textbf{MambaTG} with \textbf{MambaTG without DRR} to isolate the contribution of DRR within MambaTG. We conduct these comparisons on UCI and GoogleLocal, representing repeated- and unseen-edge-dominant settings, respectively. Within each framework, the variants with and without DRR use the same model configuration and training settings, differing only in the inclusion of DRR. On UCI, all variants additionally use historical repetition frequency and are denoted by the suffix ``-R.'' The CRAFT results in this analysis are obtained from our own experiments and may therefore differ from the originally reported results presented in Table~\ref{tab:main_results}.

\paragraph{Results.}
Figure~\ref{fig:drr_impact} shows that DRR improves both CRAFT and MambaTG on both datasets, indicating that its benefits are not specific to either architecture. Comparing the two frameworks without DRR, MambaTG consistently outperforms CRAFT, showing that its candidate-aware sequential modeling provides an advantage independently of DRR. This advantage is retained when both frameworks use DRR, with MambaTG still outperforming CRAFT-DRR on both datasets. Thus, the gains from MambaTG cannot be attributed solely to representation refinement. Instead, the results support complementary contributions from DRR and MambaTG's joint modeling of \emph{historical-sequence evolution} and \emph{candidate--history compatibility}. The consistent improvement of CRAFT with DRR further suggests that the refined representations are independently useful for candidate--history compatibility, particularly when predicting unseen edges where the source and candidate have no previous direct interaction.

\begin{figure}[h]
    \centering
    \includegraphics[width=\linewidth]{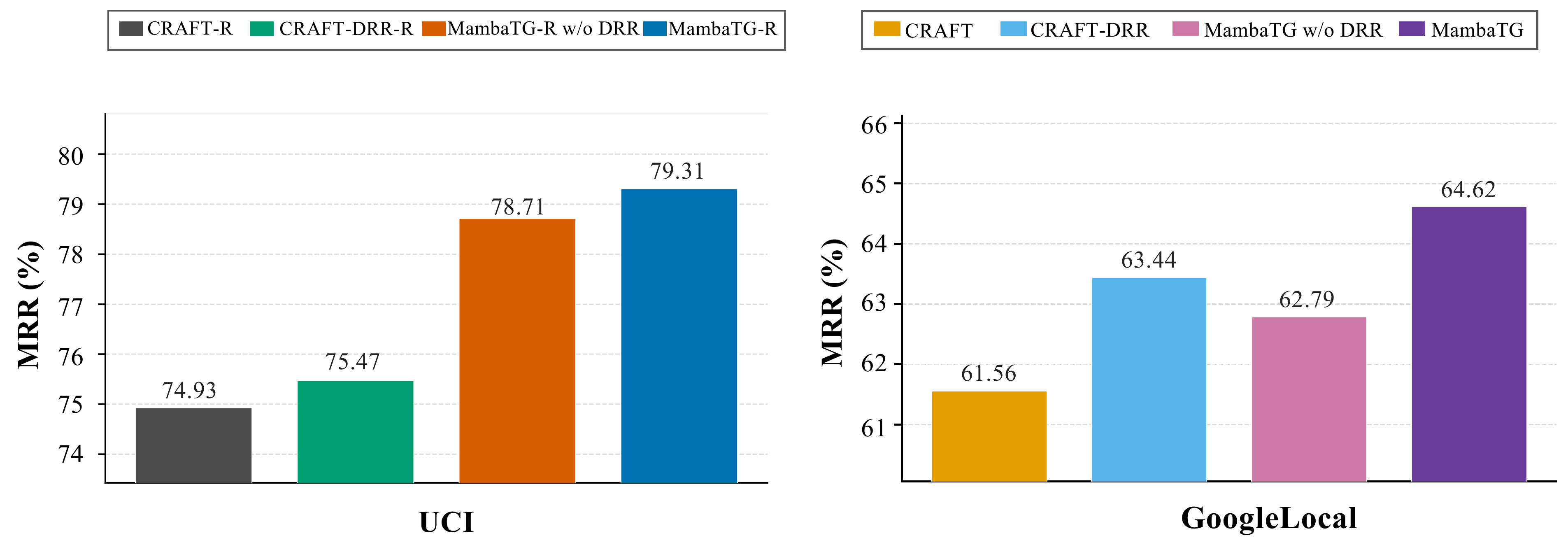}
    \caption{Impact of DRR on CRAFT and MambaTG on UCI (repeated-edge-dominant) and GoogleLocal (unseen-edge-dominant) datasets. Results are reported as MRR (\%), averaged over three random seeds.}
    \label{fig:drr_impact}
\end{figure}

\section{Effectiveness of Candidate-Aware Mamba Encoder (CAME)}
\label{app:came_impact}

CAME jointly models \emph{historical-sequence evolution} and \emph{candidate--history compatibility} by treating each candidate as a continuation of the source node's interaction sequence. To assess the role of Mamba-based sequence modeling within CAME, we compare MambaTG with GRUTG. GRUTG replaces each Mamba block with a Gated Recurrent Unit (GRU) of the same hidden dimension while retaining all other model components, training settings, and the candidate-continuation procedure. We evaluate both models on UCI and GoogleLocal, representing repeated-edge-dominant and unseen-edge-dominant settings, respectively. On UCI, both models additionally use historical repetition frequency and are denoted as GRUTG-R and MambaTG-R.

\paragraph{Results.}
As shown in Figure~\ref{fig:came_impact}, MambaTG consistently outperforms GRUTG on both datasets, indicating that Mamba provides a more effective sequential backbone for CAME. This advantage may stem from Mamba’s structured SSM state and selective recurrence, which capture different aspects of the interaction history over different timescales and selectively retain informative events. The larger gain on the unseen-edge-dominant GoogleLocal dataset further suggests that Mamba is particularly useful when direct source--candidate evidence is limited and prediction relies more heavily on patterns in the source history. Overall, these results support the use of Mamba as the sequential encoder for jointly modeling \emph{historical-sequence evolution} and \emph{candidate--history compatibility} in CAME.

\begin{figure}[h] 
\centering 
\includegraphics[width=1\linewidth]{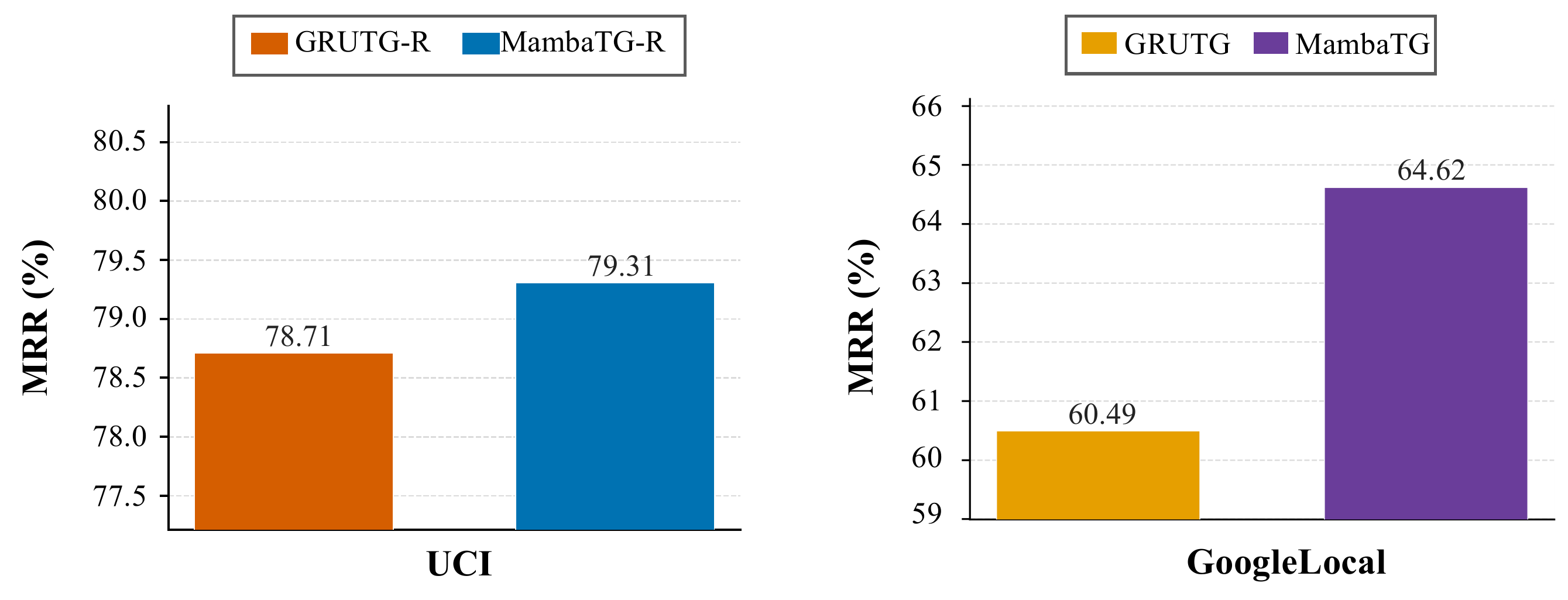} 
\caption{Effect of CAME on the repeated-edge-dominant UCI dataset and the unseen-edge-dominant GoogleLocal dataset. MambaTG is compared with GRUTG. Results are reported as MRR (\%), averaged over three random seeds.}
\label{fig:came_impact} 
\end{figure}

\section{Comparison with Recommendation Models}
\label{app:recommendation_comparison}

MambaTG targets future link prediction on temporal graphs, a general task arising in diverse applications such as recommendation, social networks, communication systems, and transportation networks. It supports both bipartite and non-bipartite graphs and covers both repeated and unseen future edges. Recommendation can be viewed as a specific application of this task, where temporal interactions typically occur between users and items in a bipartite graph.

To further examine the effectiveness of MambaTG, we compare it with two representative sequential recommendation models: the Transformer-based SASRec~\citep{kang2018selfattentive} and the Mamba-based Mamba4Rec~\citep{liu2024mamba4rec}. The comparison covers MOOC and GoogleLocal, representing repeated-edge-dominant and unseen-edge-dominant settings, respectively. As shown in Figure~\ref{fig:recommendation_comparison}, MambaTG or MambaTG-R consistently outperforms both recommendation models across both datasets. These results suggest that models designed specifically for sequential user--item recommendation may not fully capture the broader characteristics of future link prediction, particularly the distinction between repeated and unseen edges.

\begin{figure}[h]
    \centering
    \includegraphics[width=\linewidth]
    {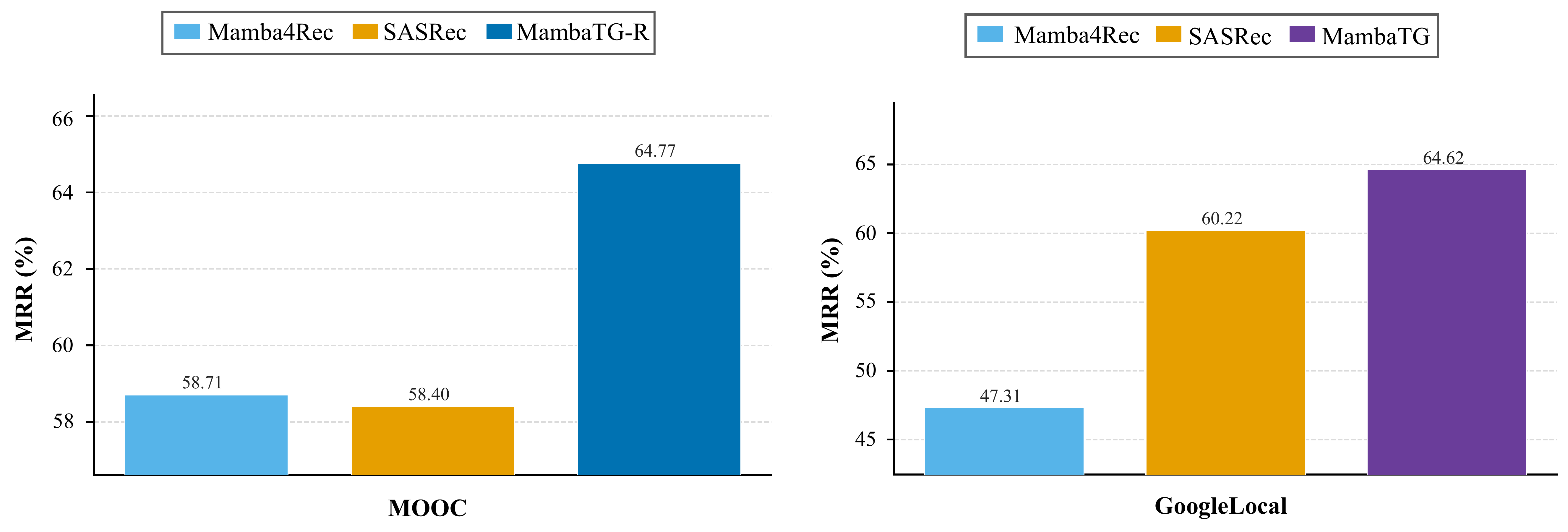}
    \caption{Comparison with sequential recommendation models on MOOC (repeated-edge-dominant) and GoogleLocal (unseen-edge-dominant) datasets. Results are reported as MRR (\%), averaged over three random seeds.}
    \label{fig:recommendation_comparison}
\end{figure}

\section{Time Complexity Analysis}
\label{app:complexity_derivations}

We analyze a batch of \(B\) source nodes, each evaluated against \(q\) candidate destinations. Let \(\ell\) denote the number of historical interactions per node, \(d\) the feature dimension, and \(\bar{d}\) the average node degree. We report only dominant terms and treat architectural hyperparameters, including the number of layers and attention heads, as well as the Mamba expansion factors and state dimensions, as constants. For TGAT, however, we explicitly account for its two-layer neighborhood expansion, which introduces two-hop histories and changes the dependence on \(\ell\).

\paragraph{Historical-neighbor extraction.}
We use neighbor extraction to refer to the temporal-index lookup operations required to locate the relevant histories in node adjacency lists. Following the common implementation in DyGLib~\citep{dygformer}, retrieving the most recent historical interactions before the query time requires \(O(\log\bar{d})\) per node. TGN~\citep{tgns}, GraphMixer~\citep{graphmixer}, DyGFormer, DyG-Mamba~\citep{li2025dyg-mambacontinuousstatespace}, and DyGMamba~\citep{ding2025dygmambaefficientlymodelinglongterm} retrieve histories for both the source and all candidate destinations, resulting in \(O(B(1+q)\log\bar{d})\). TGAT~\citep{tgat} additionally performs temporal-index lookups for up to $\ell$ first-hop neighbors in its two-layer architecture. Thus, it performs $O(B(1+q)(1+\ell))$ lookup operations, each costing $O(\log\bar{d})$, resulting in a neighbor-extraction cost of $O(B(1+q)(1+\ell)\log\bar{d})$. In contrast, CRAFT~\citep{craft} and MambaTG retrieve only the source history, reducing the cost to \(O(B\log\bar{d})\).

\paragraph{Edge-likelihood computation.}
TGN and GraphMixer process the \(\ell\) historical interactions of both the source and every candidate destination, resulting in \(O(B(1+q)\ell d^2)\). 
TGAT aggregates two-hop histories with two graph-attention layers, where \(\ell\) is the sampling size at each hop, requiring \(O(B(1+q)\ell^2d^2)\). DyGFormer applies feature projection and self-attention to each endpoint, with costs of \(O(\ell d^2)\) and \(O(\ell^2d)\), respectively. Its total complexity is therefore \(O(B(1+q)(\ell d^2+\ell^2d))\). DyG-Mamba and DyGMamba process the full history for every source--candidate pair using state-space models, resulting in \(O(Bq\ell d^2)\). CRAFT projects the source history and candidates in \(O(B(\ell+q)d^2)\), followed by candidate-to-history attention with complexity \(O(Bq\ell d)\). Its total cost is thus \(O(B(\ell+q)d^2+Bq\ell d)\). MambaTG incurs \(O(B\ell d\log\ell)\) time for DRR, whose dominant cost arises from frequency-domain refinement of each source history. It then encodes each history once in \(O(B\ell d^2)\) and evaluates all candidates using the cached states in \(O(Bqd^2)\). Therefore, its total edge-likelihood computation is \(O(B\ell d\log\ell+B(\ell+q)d^2)\).

\section{Additional Experimental Results}
\label{app:additional_results}

\subsection{long sequence efficiency}
\label{app:additional_efficiency_results}
\begin{figure}[h]
     \centering
    \includegraphics[width=1\linewidth]{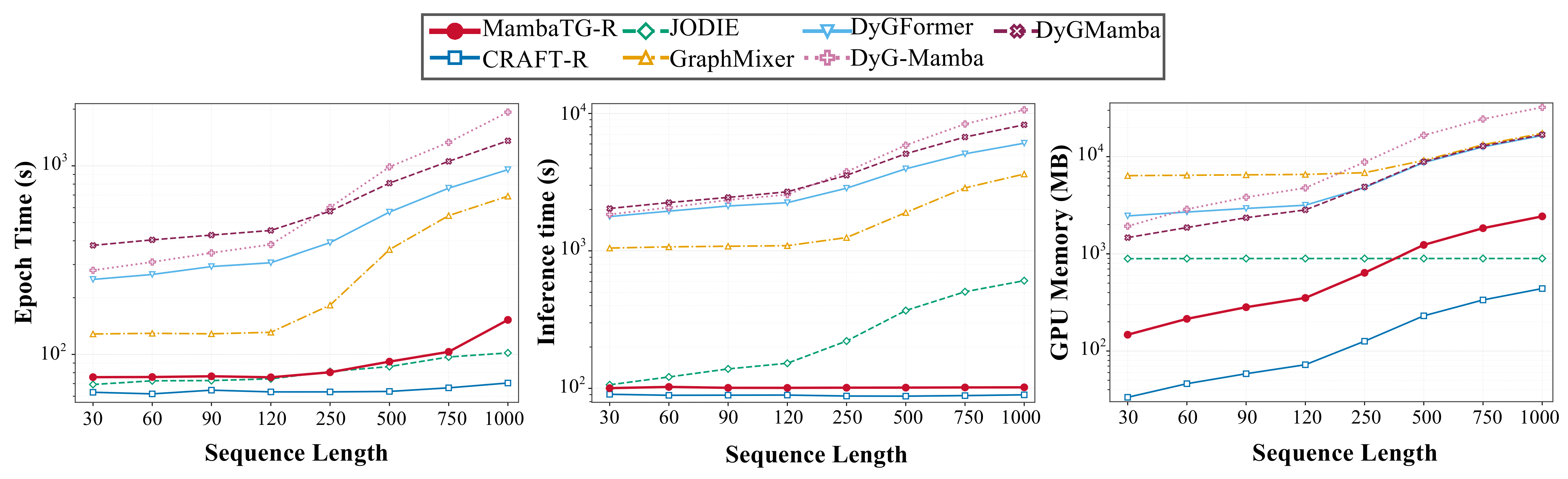}
    \caption{Computational efficiency on LastFM Dataset as the maximum source-history length \(\ell\) increases. Training time per epoch, inference time, and peak GPU memory usage.}
    \label{fig:lastfm_efficiency}
\end{figure}

We evaluate the scalability of MambaTG as the source-node history length increases on LastFM, measuring per-epoch training time, inference time, and peak GPU memory usage. Training time measures the wall-clock time of one training epoch, excluding validation and data loading before the epoch. Inference time measures the wall-clock time of the 100-negative test pass, including the operations required for each method to produce predictions. Peak GPU memory is measured as the maximum allocated GPU memory over the full run, including training and evaluation. 

Results are shown in Figure~\ref{fig:lastfm_efficiency}. (1) \emph{Training efficiency.} MambaTG scales efficiently with increasing history length and remains among the faster methods. At \(\ell=750\), it achieves up to \(12.91\times\) speedup over DyG-Mamba while remaining competitive with CRAFT and JODIE. (2) \emph{Inference efficiency.} Although inference time scales near-linearly with \(\ell\) in theory, it remains nearly constant empirically because candidate evaluation dominates wall-clock time. MambaTG caches historical states and reuses them across candidates, avoiding repeated history encoding. At \(\ell=1000\), it achieves up to \(104.54\times\) speedup over the compared baselines. (3) \emph{Memory efficiency.} MambaTG exhibits approximately linear memory growth with history length and remains substantially more memory-efficient than several baselines. Although JODIE becomes more memory-efficient for longer histories due to its fixed-size node states, MambaTG maintains a competitive memory footprint across the evaluated sequence lengths. Overall, these results demonstrate that MambaTG can effectively scale to long source histories while maintaining low computational overhead, making it well suited to large-scale recommendation scenarios where a user's extensive interaction history must be evaluated against many candidate items.

\subsection{Performance--Efficiency Trade-off}
\label{app:performance_efficiency_trade_off}

We additionally evaluate the performance--efficiency trade-off on LastFM and UCI. Figure~\ref{fig:additional_pareto_analysis} plots test MRR against per-epoch training time, with bubble size representing parameter footprint.

\begin{figure}[t]
    \centering
    \begin{minipage}[t]{0.49\linewidth}
        \centering
        \includegraphics[width=\linewidth]{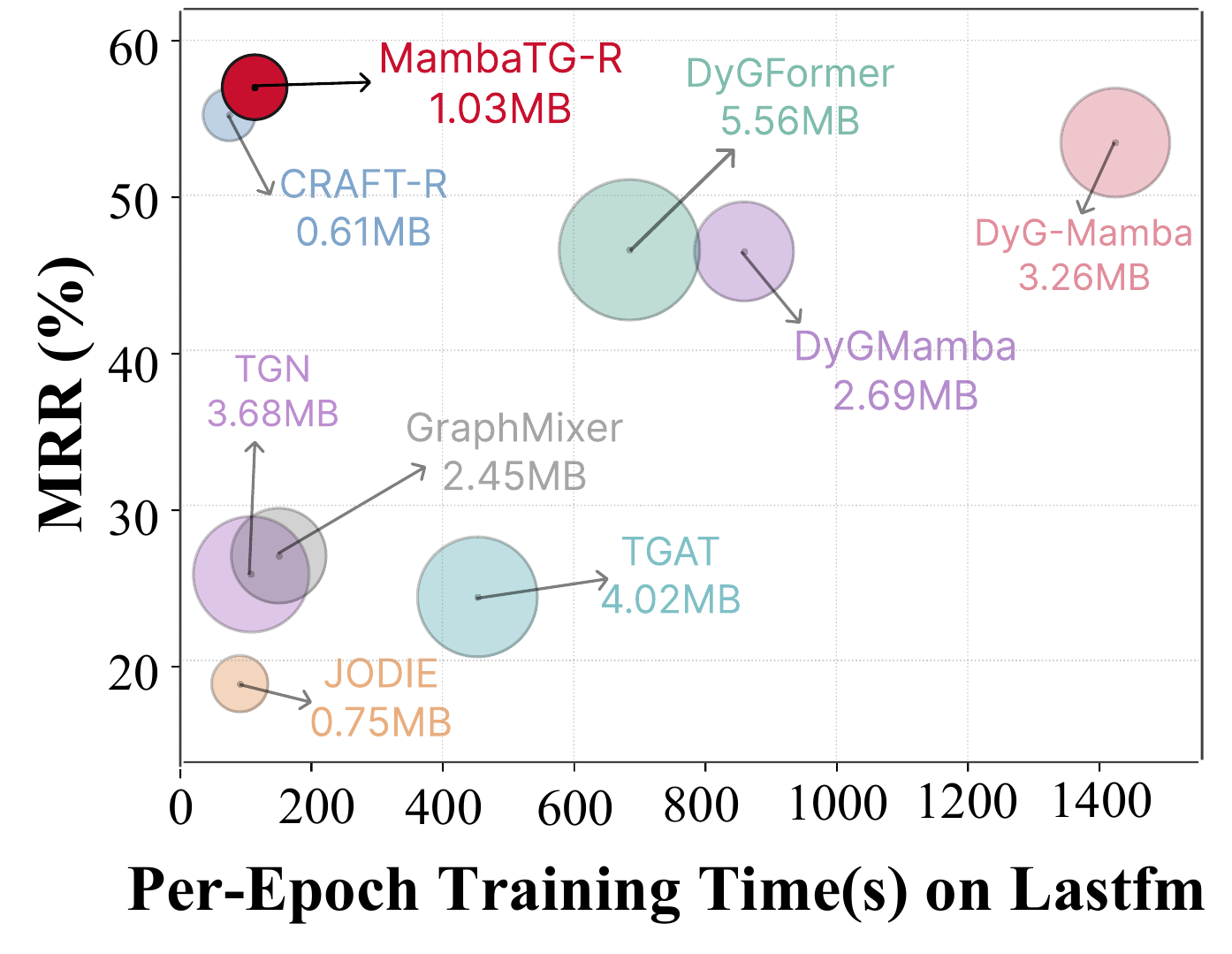}
        \textbf{(a)}
    \end{minipage}
    \hfill
    \begin{minipage}[t]{0.49\linewidth}
        \centering
        \includegraphics[width=\linewidth]{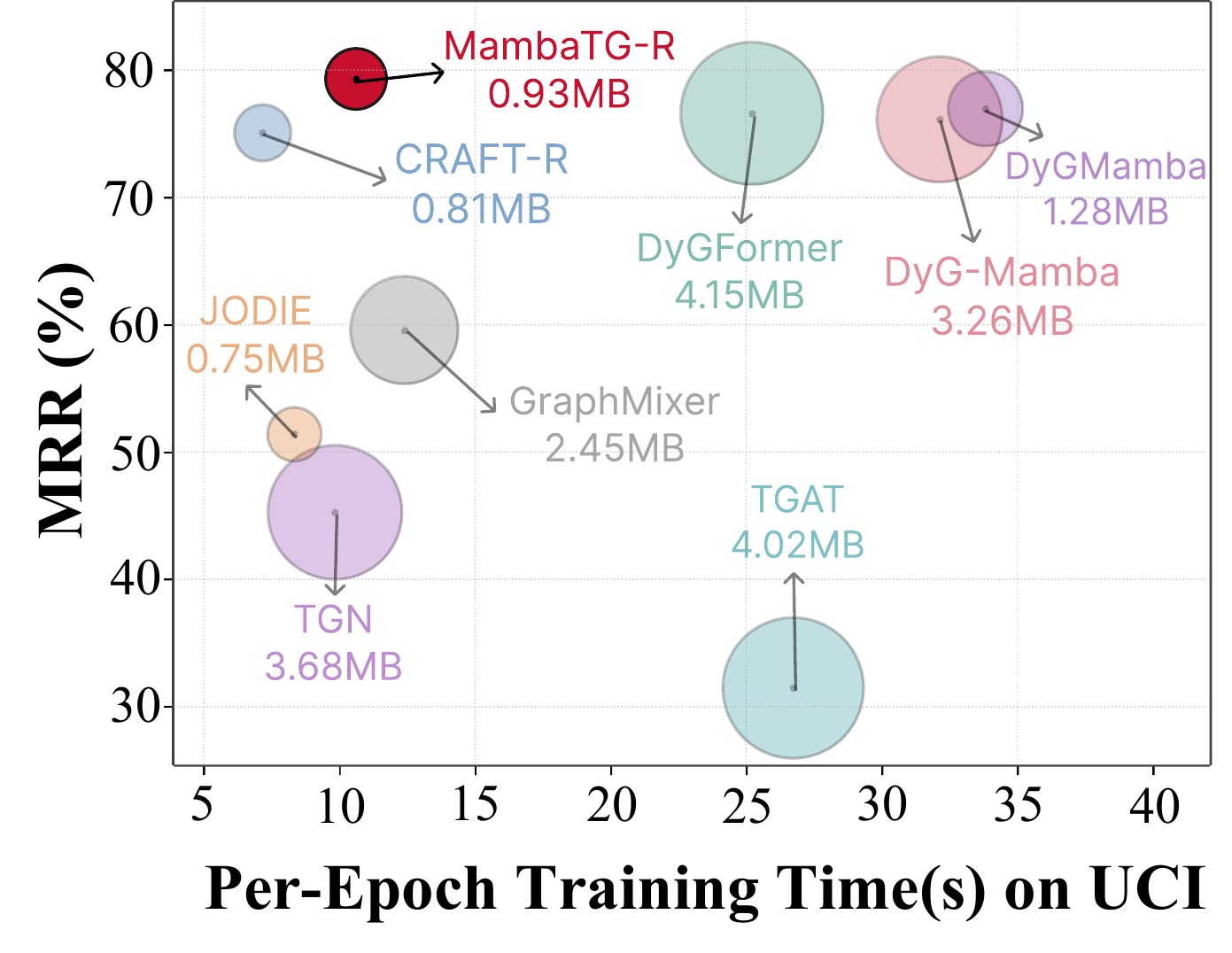}
        \textbf{(b)}
    \end{minipage}
    \caption{Performance--efficiency trade-off on (a) LastFM and (b) UCI. Test MRR is plotted against per-epoch training time, with bubble size representing parameter footprint.}
    \label{fig:additional_pareto_analysis}
\end{figure}

\subsection{Additional results on other datasets}
\label{app:additional_results_tgbl_datasets}
We further evaluate MambaTG on three datasets from the Temporal Graph Benchmark (TGB) \citep{huang2023temporalgraphbenchmarkmachine}: \texttt{tgbl-comment}, \texttt{tgbl-coin}, and \texttt{tgbl-flight}. We compare against DyGFormer~\citep{dygformer}, SGNN-HN~\citep{pan2020star}, and CRAFT~\citep{craft}. The DyGFormer results are taken from \citet{yu2023empiricalevaluationtemporalgraph}, whereas the SGNN-HN and CRAFT results are obtained from our experiments. The two prior Mamba-based baselines, DyGMamba and DyG-Mamba, are not reported because they exceed the available computational budget on all three datasets. As shown in Table~\ref{tab:additional_tgb_results}, MambaTG achieves superior performance across all three datasets. These results further demonstrate the effectiveness and generalization of MambaTG across diverse temporal graph datasets.

\begin{table}[h]
\centering
\caption{Future link prediction results on three additional TGB
datasets. Results are reported as MRR (\%), averaged over three random seeds. The \best{best} and
\second{second-best} results in each column are highlighted. Missing entries indicate experiments that exceeded computational budget.}
\label{tab:additional_tgb_results}

\begin{tabular}{lcccc}
\toprule
\multicolumn{2}{c}{\textbf{Unseen-edge-dominant}} &
\multicolumn{3}{c}{\textbf{Repeated-edge-dominant}} \\
\cmidrule(lr){1-2} \cmidrule(lr){3-5}
\textbf{Model} & \textbf{tgbl-comment} &
\textbf{Model} & \textbf{tgbl-coin} & \textbf{tgbl-flight} \\
\midrule

DyGFormer
& $67.03 \pm 0.14$
& DyGFormer
& \second{$75.17 \pm 0.38$}
& -- \\

SGNN-HN
& $40.38 \pm 0.18$
& SGNN-HN
& $66.25 \pm 0.41$
& $81.86 \pm 0.07$ \\

CRAFT
& \second{$75.69 \pm 0.75$}
& CRAFT-R
& $74.43 \pm 0.30$
& \second{$86.42 \pm 0.05$} \\

\textbf{MambaTG}
& \best{$75.93 \pm 0.21$}
& \textbf{MambaTG-R}
& \best{$75.74 \pm 0.28$}
& \best{$86.53 \pm 0.08$} \\

\bottomrule
\end{tabular}
\end{table}

\section{Discussion}
\subsection{Limitations and Future Work}
Although this work focuses on future link prediction, we have not examined whether the two signals modeled by MambaTG generalize to other temporal-graph prediction tasks. Extending MambaTG to tasks such as temporal edge classification and temporal anomaly detection, while investigating the roles of historical-sequence evolution and candidate--history compatibility in these settings, is a promising direction for future work.

\end{document}